\documentclass{article}

\usepackage{iclr2025_conference,times}

\usepackage[utf8]{inputenc} 
\usepackage[T1]{fontenc}    
\usepackage{hyperref}       
\usepackage{url}            
\usepackage{booktabs}       
\usepackage{amsfonts}       
\usepackage{nicefrac}       
\usepackage{microtype}      
\usepackage{xcolor}         
\usepackage{graphicx}
\usepackage{amsmath}
\usepackage{caption}
\usepackage{subcaption}
\usepackage{floatrow}
\usepackage{wrapfig}
\usepackage{algorithm}
\usepackage[noend]{algorithmic} 
\usepackage{tikz}
\usepackage{array}
\usepackage{paralist}

\usepackage{mathdots}
\usepackage{multirow, makecell, colortbl, hhline}
\usepackage{adjustbox} 
\usepackage{pgfplots}
\usepackage{pgfplotstable}
\usepackage{xfp}
\usepackage{xcolor}
\usepackage{pgfkeys}

\newcommand{\percentdiff}[2]{%
  \fpeval{round(abs((#2 - #1) / #1) * 100, 2)}%
}

\edef\QOurMethod{\fpeval{9.442}}

\edef\QSuccessRate{\fpeval{98}}
\edef\QRuntime{\fpeval{1.20}}

\edef\QCut{\fpeval{10.02}}

\edef\QABC{\fpeval{11.602}}

\edef\QABCOpt{\fpeval{9.467}}

\edef\QOptimal{\fpeval{9.371}}

\edef\QOurMethodB{\fpeval{8.883}}

\edef\QBoolformer{\fpeval{10.227}}

\edef\QCutB{\fpeval{9.341}}

\edef\QABCB{\fpeval{9.990}}

\edef\QABCOptB{\fpeval{8.909}}

\edef\QOptimalB{\fpeval{8.849}}

\pgfkeys{
    /Ablation/.is family, /Ablation,
    S1/.initial = {}, T1/.initial = {}, Q1/.initial = {},
    S2/.initial = {}, T2/.initial = {}, Q2/.initial = {},
    S4/.initial = {}, T4/.initial = {}, Q4/.initial = {},
    S8/.initial = {}, T8/.initial = {}, Q8/.initial = {},
    S16/.initial = {}, T16/.initial = {}, Q16/.initial = {},
    S32/.initial = {}, T32/.initial = {}, Q32/.initial = {},
    S64/.initial = {}, T64/.initial = {}, Q64/.initial = {},
    S128/.initial = {}, T128/.initial = {}, Q128/.initial = {},
    S256/.initial = {}, T256/.initial = {}, Q256/.initial = {},
    S512/.initial = {}, T512/.initial = {}, Q512/.initial = {},
}

\pgfkeys{/Ablation/S1 = 92.2} \pgfkeys{/Ablation/T1 = 0.11} \pgfkeys{/Ablation/Q1 = 9.633}
\pgfkeys{/Ablation/S2 = 97.2} \pgfkeys{/Ablation/T2 = 0.99} \pgfkeys{/Ablation/Q2 = 9.490}
\pgfkeys{/Ablation/S4 = 97.4} \pgfkeys{/Ablation/T4 = 1.93} \pgfkeys{/Ablation/Q4= 9.458}
\pgfkeys{/Ablation/S8 = 98.0} \pgfkeys{/Ablation/T8 = 3.50} \pgfkeys{/Ablation/Q8 = 9.442}
\pgfkeys{/Ablation/S16 = 98.8} \pgfkeys{/Ablation/T16 = 6.39} \pgfkeys{/Ablation/Q16 = 9.438}
\pgfkeys{/Ablation/S32 = 98.2} \pgfkeys{/Ablation/T32 = 11.85} \pgfkeys{/Ablation/Q32 = 9.409}
\pgfkeys{/Ablation/S64 = 98.4} \pgfkeys{/Ablation/T64 = 21.66} \pgfkeys{/Ablation/Q64 = 9.430}
\pgfkeys{/Ablation/S128 = 98.4} \pgfkeys{/Ablation/T128 = 53.72} \pgfkeys{/Ablation/Q128 = 9.411}
\pgfkeys{/Ablation/S256 = 98.6} \pgfkeys{/Ablation/T256 = 106.45} \pgfkeys{/Ablation/Q256 = 9.375}

\usepgfplotslibrary{groupplots}
\usetikzlibrary{patterns}
\pgfplotsset{compat=1.18}

\usetikzlibrary{shapes,arrows.meta,calc,positioning, matrix, decorations.pathreplacing,calligraphy}
\newfloatcommand{capbtabbox}{table}[][\FBwidth]

\definecolor{color1}{RGB}{255,100,100} 
\definecolor{color2}{RGB}{20,175,20} 
\definecolor{color3}{RGB}{50,50,255} 
\definecolor{color4}{RGB}{10,70,50} 
\definecolor{white}{RGB}{255,255,255} 

\newcommand{\OurMethod}{ShortCircuit}
\newcommand{\Target}{T_\star}

\title{\OurMethod{}: AlphaZero-Driven Generative Circuit Design}

\author{
Dimitrios Tsaras$^{1,2}$, Antoine Grosnit$^{1,3}$, 
    \\
    {\bfseries
    \ Lei Chen$^{1}$, Zhiyao Xie$^{2}$,  
    }
    \\
    {\bfseries
    \ Haitham Bou-Ammar\thanks{Equal supervision}$\ \ ^{1,4}$, Mingxuan Yuan\footnotemark[1]$\ \ ^{1}$
    }
    \\
    \ Noah's Ark Lab, Huawei$^{1}$,\  HKUST$^{2}$,\ TU Darmstadt$^{3}$,\  University College London$^{4}$
}

\iclrfinalcopy  

\begin{document}

\maketitle

\begin{abstract}
Chip design relies heavily on generating Boolean circuits, such as AND-Inverter Graphs (AIGs), from functional descriptions like truth tables.
This generation operation is a key process in logic synthesis, a primary chip design stage.
While recent advances in deep learning have aimed to accelerate circuit design, these efforts have mostly focused on tasks other than  synthesis, and traditional heuristic methods have plateaued.
In this paper, we introduce \OurMethod{}, a novel transformer-based architecture that leverages the structural properties of AIGs and performs efficient space exploration.
Contrary to prior approaches attempting end-to-end generation of logic circuits using deep networks, \OurMethod{} employs a two-phase process combining supervised with reinforcement learning to enhance generalization to unseen truth tables.
We also propose an AlphaZero variant to handle the double exponentially large state space and the reward sparsity, enabling the discovery of near-optimal designs.
To evaluate  the  generative performance of our model , we extract 500 truth tables from a set of 20 real-world circuits.
\OurMethod{} successfully generates AIGs for $\QSuccessRate\%$ of the
8-input
test truth tables, and outperforms
the state-of-the-art logic synthesis tool, \texttt{ABC}, by $\percentdiff{\QABC}{\QOurMethod}\%$ in terms of circuits size.
\end{abstract}

\section{Introduction} \label{sec:introduction}

The rapid proliferation of AI has triggered an unprecedented surge in computational demands, exceeding the capabilities of existing hardware and thereby becoming a major bottleneck to AI's continued growth.
Chip design plays a pivotal role in enabling the next-generation of computing systems.
However, traditional methodologies struggle to keep pace with the accelerating demands, underscoring  the need for innovative chip design approaches to accelerate the design process and discover novel architectures. 
At its core, a chip is the physical embodiment of a Boolean function, transforming binary inputs into desired outputs.
Creating these embodiments is facilitated by logic synthesis, a crucial step in chip design that converts functional descriptions into directed graphs comprising logic gates.
The resulting circuit must balance competing objectives, including power efficiency, performance, and silicon area (PPA), presenting a formidable optimization challenge. 
In this paper, we investigate the use of Machine Learning (ML) to generate optimized digital circuits directly from Boolean logic specifications, offering a fresh perspective on the chip design process.

Truth tables provide a complete and unambiguous way of representing a Boolean function by exhaustively enumerating the output values for all possible binary input combinations. 
As such, we use truth tables as the input Boolean logic description for our problem.
The output of our approach are directed graphs (DAGs) in the form of an AND-Inverter Graphs (AIGs), a widely used data structure in Electronic Design Automation (EDA)~\citep{Mishchenko2006ScalableLS, Wolf2013YosysAFV}.
AIGs consist of 2-input AND-gates as nodes, connected with normal or inverted wires, offering a simple, scalable, and universal intermediate representation for various EDA applications.
Their popularity in the industry stems from their ability to efficiently represent complex Boolean functions, making them an ideal choice for our ML-based circuit generation approach.

Recent efforts attempt to accelerate chip production by leveraging the development of ML methods at different steps of EDA~\citep{ Huang2021SurveyML4EDA, Gubbi2022SurveyML4EDA}
, notably for placement~\citep{Ward2012KeepIS}, 
routing~\citep{Alawieh2020Routing}, 
and for logic synthesis~\citep{Tu2024AI4LogicSynthesis}. 
Rather than directly tackling the graph generation problem,  most ML methods for logic synthesis focus on the optimization of synthesis recipes,
which are sequences of operators acting on a logic graph to modify its structure while preserving the associated boolean function.
More recently, deep-generative methods emerged aiming to generate logic graphs rather than sequences of graph operators~\cite{ascoli2024boolformer, Li2024CircuitTransformer, dong2023cktgnn}. 
The generative approach offers higher potential as it offers more flexibility than working with a fixed set of operators; on the other hand, 
this approach involves exploring a much larger space, due to the double exponential growth of the search space with the number of Boolean function inputs.

Such a vast search space renders traditional ML methods ineffective for optimal AIG generation.
However, recent advances have demonstrated that, with tailored model architectures and exploration-exploitation-aware training protocols, remarkable performance can be achieved even in tasks involving such large spaces.
Indeed, we can attribute the success of methods like  AlphaGO~\cite{Silver2016AlphaGO}, AlphaZero~\cite{Silver2017AlphaZero}, AlphaFold~\cite{Jumper2021AlphaFold} and even the emergence of large language models~\cite{Devlin2019BERTPO, Kaplan2020ScalingLF} 
to the development of custom model architectures capturing structural properties of board games, proteins, or language.
Moreover, the training of these models  either leverages naturally abundant data or employs specific data-augmentation and exploration-exploitation strategies to improve their performance.

In this work, we propose \OurMethod{}, a new transformer-based architecture structurally adapted for generating AND-Inverter graphs. 
Our transformer takes logic nodes represented as truth tables as input, and each forward pass predicts the next AND node to create in order to realize a target truth table. 
Moreover, we utilize an AlphaZero policy variant to effectively navigate the large state space and discover more compact designs. 

We summarize our contributions as follows. \textbf{i)} We formally define the challenging problem of generating AIGs from target truth tables, characterized by a doubly exponential state space and a quadratically expanding action space. 
\textbf{ii)} We introduce \OurMethod{}, a novel AIG-aware transformer-based architecture, enabling effective exploration of this vast search space. 
\textbf{iii)} We propose a two-step training approach, combining supervised learning and reinforcement learning, to efficiently learn generalizable patterns and prune the search space, improving solution quality and scalability.
Finally, \textbf{iv)} we empirically demonstrate the effectiveness of \OurMethod{} by producing circuits $\percentdiff{\QABC}{\QOurMethod}\%$ smaller compared to the state-of-the-art logic synthesis, tool \texttt{ABC}~\citep{mishchenko2007abc}, and showcase the potential of ML methods to revitalize the field of logic synthesis with a fresh perspective.


We organize the rest of the paper by introducing  the necessary concepts and related works in section~\ref{sec:background}, which provides a grounding for the formal problem formulation that we present in section~\ref{sec:problem}. 
We then detail our proposed approach, notably our model architecture~\ref{sec:method} and our tailored training procedure in section~\ref{sec:training}.
We finally present in section~\ref{sec:exp_evaluation} an empirical evaluation of our method's performance.

\section{Background} \label{sec:background}

A digital circuit cascades logic gates to realize a Boolean function $f:\{0,1\}^n \rightarrow \{0,1\}^m$, mapping a Boolean vector of size $n$ to a Boolean vector of size $m$.
An And-Inverter Graph (AIG) is a Directed Acyclic Graph (DAG) that is commonly used to represent a Boolean function at the early stage of the chip design process, due to its simplicity and ubiquity.

\subsection{AND-Inverter Graphs}
An AIG is composed of three types of nodes, (1) primary
inputs, which we also refer to as inputs, (2) primary outputs, which we call the outputs,
and (3) 2-input AND-nodes representing the logic gate AND.
In this work, we focus on the generation of single output AIGs that represent Boolean functions of the form $f:\{0,1\}^n \rightarrow \{0,1\}$ 
as they play an important role in logic synthesis.
Fig.~\ref{fig:AIG} illustrates the structure of an AIG with $n = 3$  inputs, where $\{I_k\}_{1\leq k \leq 3}$ represent input nodes, $\wedge_4, \wedge_5$ are AND gates, and $O$ is the output.
Edge orientation indicates the direction of the Boolean signal propagation from one node (called fanin) to another (called fanout). 
Moreover, the two types of edges, plain and dashed, in Fig.~\ref{fig:AIG} indicate that the Boolean signal can be inverted when going from a fanin to a fanout.
The primary output is always connected to a single AND node with a direct  or an inverter link.
As AIGs only contain AND operations and Boolean inversions, we can map them to a canonical form (CNF). 
For instance, the CNF of the AIG in Fig. \ref{fig:AIG} naturally derived  from its topology is $O=\neg(\neg(I_1\wedge I_2)\wedge  I_3)$, and we can conversely easily go from a CNF to an AIG.
Moreover, applying equivalence-preserving operations to the CNF produces new CNFs that still encode the same Boolean function.
Similarly, topologically distinct AIGs can realize the same function, and to compare the quality of two AIGs a primary criterion is to compare their sizes, measured by the number of gates they contain~\citep{Mishchenko2006ScalableLS}.
Smaller AIGs are generally preferred as they simplify subsequent tasks such as placement and routing, and lead to more efficient circuits.

\begin{figure}[!t]
\begin{floatrow}
\ffigbox{%
  \resizebox{0.45\textwidth}{!}{\begin{tikzpicture}[scale=1.5, every node/.style={scale=1.2}]
    \tikzset{
        input/.style={circle, draw, minimum size=10mm, inner sep=0pt, line width=1.2pt},
        and/.style={circle, draw, fill=gray!30, minimum size=10mm, inner sep=0pt, line width=1.2pt},
        output/.style={circle, draw, fill=gray!60, minimum size=10mm, inner sep=0pt, line width=1.2pt},
        arrownode/.style={circle, draw, minimum size=6mm, inner sep=0pt},
        legendarrownode/.style={circle, draw, minimum size=10mm, inner sep=0pt},
        normal edge/.style={->,>=latex, line width=1.2pt},
        negated edge/.style={->,>=latex, dashed, line width=1.2pt},
        node distance=2cm and 2cm
    }

    \node[input] (I1) at (0, 0) {$I_1$};
    \node[input, right=of I1] (I2) {$I_2$};
    \node[input, right=of I2] (I3) {$I_3$};
    \node[and, above=of $(I1.north)!0.5!(I2.north)$, yshift=-.4cm] (A1) {$\land_4$};
    \node[and, above=of $(A1.north)!0.35!(I3.north)$, yshift=-.1cm] (A2) {$\land_5$};
    \node[output, above=of A2, yshift=-.6cm] (O) {$O$};

    \draw[normal edge] (I1) -- (A1);
    \draw[normal edge] (I2) -- (A1);
    \draw[negated edge] (A1) -- (A2);
    \draw[negated edge] (A2) -- (O);
    \draw[normal edge] (I3) -- (A2);

    \node[right=.3cm of I1, yshift=-.3cm] (I1 val) {
            \begin{tabular}{|p{.05cm}|} \hline
                0 \\ \hline 
                1 \\ \hline 
                0 \\ \hline 
                1 \\ \hline
                0 \\ \hline 
                1 \\ \hline 
                0 \\ \hline 
                1 \\ \hline
            \end{tabular}
        };

    \node[right=0.2cm of I2,  yshift=-.3cm] (I2 val) {
            \begin{tabular}{|p{.05cm}|} \hline
                0 \\ \hline 
                0 \\ \hline 
                1 \\ \hline 
                1 \\ \hline
                0 \\ \hline 
                0 \\ \hline 
                1 \\ \hline 
                1 \\ \hline
            \end{tabular}
        };

    \node[right=.2cm of I3,  yshift=-.3cm] (I3 val) {
            \begin{tabular}{|p{.05cm}|} \hline
                0 \\ \hline 
                0 \\ \hline 
                0 \\ \hline 
                0 \\ \hline
                1 \\ \hline 
                1 \\ \hline 
                1 \\ \hline 
                1 \\ \hline
            \end{tabular}
        };

    \node[left=0.1cm of O, yshift=0cm] (O val) {
        \begin{tabular}{|p{.05cm}|} \hline
                1 \\ \hline 
                1 \\ \hline 
                1 \\ \hline 
                1 \\ \hline
                0 \\ \hline 
                0 \\ \hline 
                0 \\ \hline 
                1 \\ \hline
            \end{tabular}
    };
    \node[left=0.1cm of A1, yshift=.75cm] (A1 val) {
        \begin{tabular}{|p{.05cm}|} \hline
                0 \\ \hline 
                0 \\ \hline 
                0 \\ \hline 
                1 \\ \hline
                0 \\ \hline 
                0 \\ \hline 
                0 \\ \hline 
                1 \\ \hline
            \end{tabular}
    };
    \node[right=0.5cm of A2, yshift=0.25cm] (A2 val) {
       \begin{tabular}{|p{.05cm}|} \hline
                0 \\ \hline 
                0 \\ \hline 
                0 \\ \hline 
                0 \\ \hline
                1 \\ \hline 
                1 \\ \hline 
                1 \\ \hline 
                0 \\ \hline
            \end{tabular}
    };

    \node[output,  right=4cm of O val, yshift=1.cm] (legendO) {$O$};
    \node[right=0.2cm of legendO] {Output Node};
    \node[and, below=0.3cm of legendO] (legendA) {$\land$};
    \node[right=0.2cm of legendA] {AND Node};
    \node[input, below=0.3cm of legendA] (legendI) {$I$};
    \node[right=0.2cm of legendI] {Input Node};
    
    \node[arrownode, draw=none, below=0.1cm of legendI] (legendNormal) {};
    \node[legendarrownode, draw=none] (legendTextPlace) at (legendNormal)  {};
    \node[right=0.2cm of legendTextPlace] {Normal Edge};
    \draw[normal edge] (legendNormal.west) -- (legendNormal.east);

    \node[arrownode, draw=none, below=0.1cm of legendNormal] (legendNeg) {};
    \node[legendarrownode, draw=none] (legendNegTextPlace) at (legendNeg)  {};
    \node[right=0.2cm of legendNegTextPlace]  (legendNegLabel) {Negated Edge};
    \draw[negated edge] (legendNeg.west) -- (legendNeg.east);

    \draw[black,thick,  line width=1.2pt] ($(legendO.north west)+(-0.3,0.2)$)  rectangle ($(legendNegLabel.south east)+(0.3,-0.2)$);

    \node[left=0.25cm of I1] (Inputs) {\large{Inputs:}};
    \draw[black,thick,dotted,  line width=1.2pt] ($(Inputs.north west)+(-0.2,0.25)$)  rectangle ($(I3.south east)+(0.9,-0.25)$);

    \node[left=1.25cm of O] (Output) {\large{Output:}};
    \draw[black,thick,dotted, line width=1.2pt] ($(Output.north west)+(-0.2,0.35)$)  rectangle ($(O.south east)+(0.4,-0.25)$);

\end{tikzpicture}}
}{%
  \caption{Representation of an AIG, showing the truth table associated to each node.}%
  \label{fig:AIG}
}
\capbtabbox{%
  \begin{tabular}{|p{.25cm}p{.25cm}p{.25cm} || p{.25cm}p{.25cm}p{.25cm}p{.25cm}p{.25cm}p{.25cm}|}
  \hline
     $I_3$ & $I_2$ & $I_1$ & $I_1$ & $I_2$& $I_3$& $\wedge_4$ & $\wedge_5$ & O  \\ \hline
     0 &  0 &  0&  0 &  0 &  0  &  0 &  0& 1\\
     0 &  0&  1 &  1  &  0&  0  &  0 &  0& 1\\
     0&  1&  0 &  0  &  1&  0 &  0 &  0& 1\\
     0&  1 &  1 &  1  &  1 &  0 &  1 &  0& 1\\
     1&  0 &  0 &  0  &  0 &  1 &  0 &  1& 0\\
     1&  0&  1 &  1  &  0 &  1 &  0&  1 & 0\\
     1 &  1&  0 &  0  &  1 &  1  &  0 &  1& 0\\
     1 &  1 &  1 &  1 &  1 &  1 &  1 &  0& 1\\ \hline
    \end{tabular}
}{
  \caption{Truth table of each node appearing in the AIG from Fig.~\ref{fig:AIG}}.
  \label{tab:truth-table}
}
\end{floatrow}
\end{figure}

\subsection{Truth Tables}

As for any other logical graphs, we can capture the behavior of an AIG by propagating Boolean values from its primary inputs to its primary outputs and applying the logical operations encountered on the directed paths. 
Considering again the exemplar AIG from Fig.~\ref{fig:AIG}, if $I_1 = 1$, $I_2 = 1$, and $I_3 = 0$, propagating the Boolean signals we can verify   that the AIG output at $O$ is $1$.
By enumerating all possible input combinations and recording the corresponding output values for each gate, we can build the AIG's full truth table,  as shown in Table~\ref{tab:truth-table}.
Each row corresponds to the values of the AIG nodes for a specific set of entries $(I_1, I_2, I_3) = (i_1, i_2, i_3) \in \{0, 1\}^3$ displayed on the left part of the table.
We can extract from this representation a binary vector of size $2^n$ for each AIG node.
For instance, the "truth table" vector representation of node $\wedge_4$ is $(0\ 0 \ 0\ 1\ 0\ 0\ 0\ 1)^\top$, and the primary output one is $(1\ 1 \ 1\ 1\ 0\ 0\ 0\ 1)^\top$. 
After discussing related works in more details, we will explain in the next section how our method utilizes this rich vector representation to perform AIG generation.

\subsection{Related works} \label{sec:related-works}

We discuss heuristics for AIG generation and ML methods for logic operator sequence optimization in Appendix~\ref{app:add-related-works}, and focus here on the deep learning approaches tailored to logic graph generation.

\paragraph{Learning to Generate One Circuit at a Time}
A first approach to generate Boolean networks with deep neural networks consists of substituting the gates and wires of a logic circuit by learnable nodes and connections to form a neural network~\citep{Belcak2022NeuralComb, zimmer2023differentiable, hillier2023learning}. 
The network parameters are learnt by minimizing  the error made by forward passes compared to the target binary output.
On the one hand, this method allows to cope with larger number of primary inputs as, instead of learning an entire family of logic graphs (e.g., the 8-input AIGs), it is specialized on one particular target truth table. On the other hand, it requires to train a new neural network for each target truth table, representing a significant runtime bottleneck.
Therefore, several works inspired by the development of foundational generative models have followed another direction, consisting in learning the synthesis process itself with a deep neural network. 

\paragraph{Learning Circuit Synthesis using Deep Learning}
While \citet{Roy2021PrefixRL} use a CNN backbone to generate  prefix circuits by adding or deleting nodes in a $N \times N$ grid,  \citet{li2024layerdag}  employ an auto-regressive diffusion model to generate DAG for high-level synthesis stage 
, and \citet{dong2023cktgnn} design a two-level GNN architecture to synthesize analog circuits that work with non-binary signals.
Closer to our work are Boolformer~\citep{ascoli2024boolformer} and Circuit Transformer (CT)~\citep{Li2024CircuitTransformer}, both tackling digital network synthesis with an auto-regressive transformer-based architecture.
They train their policies with a supervised training phase, predicting the next element of a logic graph given a target truth table, and doing inference through beam-search or MCTS simulations.
Contrary to our work, Boolformer and CT use a symbolic representation of the Boolean formula associated to a logic graph that is encoded via depth-first search. 
Therefore instead of representing already built nodes by their truth tables and predicting the next node to add to the graph, they tokenize the symbols and let their model generate the next symbol of the logic formula (a logical operation, an input or an output), which requires more forward passes than for our method to produce a similar AIG.
Besides, given a target truth table $\Target{}$, Boolformer directly takes as input the boolean representation of $\Target{}$  as we do, while \citet{Li2024CircuitTransformer} pass a (non-optimal)  AIG realizing $\Target{}$ 
to the CT to generate an improved logic network. 





\section{Problem Definition} \label{sec:problem}


Truth tables encompass the results for all possible input values, but do not provide sufficient structural information to derive an AIG with the corresponding output representation. 
Heuristics that generate a Boolean expression, or equivalently an AIG, from a truth table generally lead to exponentially large solutions that need further refinement. 
Consequently, solving this AIG generation problem would significantly impact digital circuit design.

\textbf{Problem Definition:} Given a target truth table $\Target{}\in\{0,1\}^n$, construct an AIG with the minimum number of nodes, such that its output node $O$ has a truth table $T_O$ that matches $\Target{}$. 

Note that the size of a truth table associated with an $n$-input AIG is $2^n$,  thus, the set containing all truth tables of size $2^n$ has a cardinality of $2^{2^n}$, which corresponds to a magnitude of $10^{19}$ 
for the space of truth tables that a 6-input AIG can represent.
Exploring such a large space efficiently requires developing specific models and training techniques.
These must exploit the structural properties of the problem while managing the exploration-exploitation trade-off inherent in such scenarios.

\subsection{State Representation \& Notations}

We formally define an AIG with $n$ inputs as a graph $\mathcal{G}=(V,E)$, where $V$ and $E$ represent the node and edge sets. 
To capture the dynamic nature of AIG construction, we introduce a temporal parameter, $t$, which simultaneously represents the current time step and the number of AND-nodes in the graph, denoted as $\mathcal{G}_t=(V_t,E_t)$.
This allows us to model the evolution of the AIG over time, with the graph growing as new AND-nodes are added.
We assign a unique integer ID to each node in the graph, regardless of its type; thus, at time $t$ the node set is $V_t=\{I_1,\dots,I_n,\wedge_{n+1},\dots, \wedge_{n+t}\}$, or with a node-type agnostic notation, $V_t = \{v_1, \dots, v_{n+t}\}$. 
Following this notation, the graph $\mathcal{G}_0$ represents an AIG containing only input nodes, i.e., with $V_0=\{I_1,I_2,...,I_n\}$. 

In our generative process, we encode the state corresponding to AIG $\mathcal{G}_t$ as a 3-tuple $s_t=(\mathcal{T}_t, \Target{}, \mathcal{A}_t)$. Here, $\mathcal{T}_t=\{T_1, T_2,...,T_{n+t}\}$ is the set of truth tables associated with the current nodes, $\Target{}$ is the target truth table, and $\mathcal{A}_t=\{a_{n+1}, a_{n+2},...,a_{n+t}\}$ is the ordered set of actions performed so far.
Each action generates a new AND-node by connecting two existing nodes in one of four possible configurations: $(v_i, v_j)$, $(v_i, \neg v_j)$, $(\neg v_i, v_j)$, and  $(\neg v_i, \neg v_j)$.
Our goal is to perform a series of $N$ actions, transforming $\mathcal{G}_0$ into a terminal AIG $\mathcal{G}_N$ such that the truth table of the last generated AND-node, $T_{n+N}$, matches or is the negation of the target truth table $\Target{}$. 
Note that our environment is stateful, as its history influences future decisions.
Furthermore, our environment poses an additional challenge due to its action space expanding quadratically with each new node we add.

\section{Model Architecture} \label{sec:method}

\begin{figure}[!t]
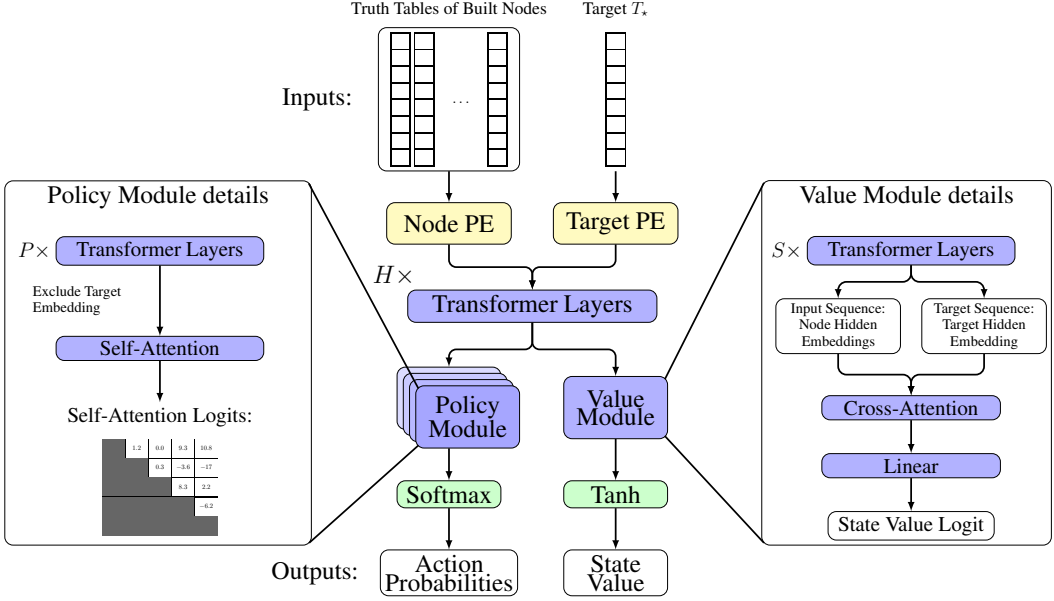

\centering
    \resizebox{\textwidth}{!}{\begin{tikzpicture}
    Styles
    \tikzset{
        round node/.style={draw, rounded corners=0.2cm},
        pol module/.style={round node, minimum width=2.5cm, minimum height=1.5cm},
        value module/.style={round node, minimum width=2.5cm, minimum height=1.5cm},
        normal connection/.style={->, >=latex, line width=1.2pt},
    }

\node [round node, fill=yellow!30, minimum width=3cm, minimum height=1.cm] (node_pe) at (0,0) {\LARGE Node PE};
\node [round node, fill=yellow!30, right=of node_pe, minimum width=3cm, minimum height=1.cm] (target_pe) {\LARGE Target PE};
\node [round node, fill=blue!30, below=1.6cm of  $(node_pe)!0.5!(target_pe)$,  minimum width=6cm] (arch_transf_layers) {\LARGE Transformer Layers};
\node [draw=none, above left=0cm of arch_transf_layers, xshift=.3cm]  {\LARGE $H \times$};

\node [pol module, fill=blue!14, below=3.4cm of node_pe, xshift=0cm, yshift=.45cm]  (pol_module_1) {};

\node [pol module, fill=blue!21, below=3.4cm of node_pe, xshift=.15cm, yshift=.30cm]  (pol_module_2){};

\node [pol module, fill=blue!28, below=3.4cm of node_pe, xshift=.3cm, yshift=.15cm] (pol_module_3) {};

\node [pol module, fill=blue!35, below=3.4cm of node_pe, xshift=.45cm, yshift=0cm,align=center] (pol_module_4) {\LARGE Policy\\ \LARGE Module};
\node [value module, fill=blue!30, below=3.4cm of target_pe, yshift=.225cm, align=center]  (value_module) {\LARGE Value\\ \LARGE Module};


\node [round node, fill=green!20, below=of value_module, minimum width=2.5cm] (tanh)  {\LARGE Tanh};
\node [round node, fill=green!20, minimum width=2.5cm] at ($(pol_module_1 |- tanh)$) (softmax)   {\LARGE Softmax} ;

\node [round node, below=of softmax, minimum width=2.5cm, align=center] (action_proba) {\LARGE Action\\ \LARGE Probabilities};
\node [draw=none, left=.4cm of action_proba]  {\LARGE Outputs:};

\node [round node, below=of tanh, minimum width=2.5cm, align=center] (state_val) {\LARGE State\\ \LARGE Value};

\draw[normal connection, rounded corners=5pt] let \p1 = (node_pe.south), \p2 = (arch_transf_layers.north) in (node_pe.south) -- ($(\x1, \y1 / 2 + \y2 / 2)$) -- ($(\x2, \y1 / 2 + \y2 / 2)$) --  (arch_transf_layers.north);
\draw[normal connection, rounded corners=5pt] let \p1 = (target_pe.south), \p2 = (arch_transf_layers.north) in (target_pe.south) -- ($(\x1, \y1 / 2 + \y2 / 2)$) -- ($(\x2, \y1 / 2 + \y2 / 2)$) --  (arch_transf_layers.north);

\draw[normal connection, rounded corners=5pt] let \p1 = (arch_transf_layers.south), \p2 = (value_module.north), \p3 = (pol_module_1.north) in (arch_transf_layers.south) -- ($(\x1, \y1 / 2 + \y2 / 2)$) -- ($(\x3, \y1 / 2 + \y2 / 2)$) --  (pol_module_1.north);
\draw[normal connection, rounded corners=5pt] let \p1 = (arch_transf_layers.south), \p2 = (value_module.north) in (arch_transf_layers.south) -- ($(\x1, \y1 / 2 + \y2 / 2)$) -- ($(\x2, \y1 / 2 + \y2 / 2)$) --  (value_module.north);

\draw[normal connection] let \p1 = (pol_module_1.south), \p2 = (pol_module_4.south)
           in ($(pol_module_1.south)+(0, \y2 - \y1)$) -- (softmax);
\draw[normal connection] (value_module) -- (tanh);

\draw[normal connection] (softmax) -- (action_proba);
\draw[normal connection] (tanh) -- (state_val);

\node[round node, above=.75cm of node_pe, minimum width=3.4cm] (truth_tables) {
$\begin{array}{|m{.1cm}|}  \hline
\rule{0pt}{.1ex}  \\ \hline
\rule{0pt}{1ex}  \\ \hline
\rule{0pt}{1ex}  \\ \hline
\rule{0pt}{1ex}  \\ \hline
\rule{0pt}{1ex}  \\ \hline
\rule{0pt}{1ex}  \\ \hline
\rule{0pt}{1ex}  \\ \hline
\rule{0pt}{1ex}  \\ \hline
\end{array}$
$\begin{array}{|m{.1cm}|}  \hline
\rule{0pt}{.1ex}  \\ \hline
\rule{0pt}{1ex}  \\ \hline
\rule{0pt}{1ex}  \\ \hline
\rule{0pt}{1ex}  \\ \hline
\rule{0pt}{1ex}  \\ \hline
\rule{0pt}{1ex}  \\ \hline
\rule{0pt}{1ex}  \\ \hline
\rule{0pt}{1ex}  \\ \hline
\end{array}$
$\quad \dots\quad $
$\begin{array}{|m{.1cm}|}  \hline
\rule{0pt}{.1ex}  \\ \hline
\rule{0pt}{1ex}  \\ \hline
\rule{0pt}{1ex}  \\ \hline
\rule{0pt}{1ex}  \\ \hline
\rule{0pt}{1ex}  \\ \hline
\rule{0pt}{1ex}  \\ \hline
\rule{0pt}{1ex}  \\ \hline
\rule{0pt}{1ex}  \\ \hline
\end{array}$
};

\node[draw=none, above=.2cm of truth_tables, align=center] {\large Truth Tables of \large Built Nodes};
\node[draw=none, left=.5cm of truth_tables, align=center] {\LARGE Inputs:};

\node[draw=none, above=.75cm of target_pe] (target_truth_tables) {
$\begin{array}{|m{.1cm}|}  \hline
\rule{0pt}{.1ex}  \\ \hline
\rule{0pt}{1ex}  \\ \hline
\rule{0pt}{1ex}  \\ \hline
\rule{0pt}{1ex}  \\ \hline
\rule{0pt}{1ex}  \\ \hline
\rule{0pt}{1ex}  \\ \hline
\rule{0pt}{1ex}  \\ \hline
\rule{0pt}{1ex}  \\ \hline
\end{array}$};
\node[draw=none, above=.11cm of target_truth_tables, align=center] {\large Target $\Target{}$};

\draw[normal connection] (truth_tables) -- (node_pe);
\draw[normal connection] (target_truth_tables) -- (target_pe);

\node[round node, left=of pol_module_4, xshift=-1.5cm, yshift=1.3cm, align=center] (pol_module_details) {\LARGE Policy Module details \\ \\ \\ \input{Figures/policy-module}};
\path let \p1 = ($(arch_transf_layers.south) - (pol_module_details.south)$) in node[round node, align=center,anchor=south] (value_module_details) at ($(arch_transf_layers.south) + (\x1, -\y1)$) {\LARGE Value Module details \\ \\ \\ \input{Figures/value-module}};

\draw[line width=1.2pt] ($(value_module.north east) + (-0.07, -0.07)$) -- ($(value_module_details.north west) + (.06, -.06)$);
\draw[line width=1.2pt] ($(value_module.south east)  + (-0.07, 0.07)$) -- ($(value_module_details.south west) + (0.06, 0.06)$);

\draw[line width=1.2pt] ($(pol_module_4.north west) + (0.07, -0.07)$) -- ($(pol_module_details.north east) + (-.07, -.07)$);
\draw[line width=1.2pt] ($(pol_module_4.south west)  + (0.07, 0.07)$) -- ($(pol_module_details.south east) + (-0.06, 0.06)$);

\end{tikzpicture}}
    \caption{\OurMethod{} model takes as inputs a target truth table $\Target{}$ and the truth tables of the already built nodes. It first appends a type-dependent positional encoding before going through several transformer layers. Then, the model is split into two heads respectively outputting a probability distribution over the next possible actions (policy module on the left), and a value reflecting the quality of the current inputs (value module on the right).}
    \label{fig:Model}
\end{figure}

We propose an iterative approach to AIG construction, where we gradually build the circuit, starting with $\mathcal{G}_0$ and letting our model decide at each step which AND-node to add,  aiming at realizing a given target truth table $\Target{}$.
To generate the next gate, the model takes the set of existing nodes as input, and it outputs a probability distribution over the set of AND-nodes that can be built by combining any pair of already existing nodes, taking edge types into account.
Formally, let $|V_t|$ denote the number of nodes in the current state of the graph, then the action space is a $4\times  |V_t| \times |V_t|$ tensor, where each $|V_t|\times |V_t|$ slice corresponds to a connection type. 
Specifically, the cell $(i, j)$ in a given slice indicates the probability of connecting node $v_i$ with $v_j$, and the slice index $\epsilon \in \{1, 2, 3, 4\}$, corresponds to a combination of specific edge types: $(v_i,v_j)$, $(\neg v_i,v_j)$, $(v_i,\neg v_j)$, or $(\neg v_i,\neg v_j)$.
Therefore, we can sample a triplet $(\epsilon, i, j)$ following the distribution given by this $4\times  |V_t| \times |V_t|$ tensor and add the corresponding node to the graph. 
The process ends when the truth table of the sampled node matches either the target one $\Target{}$ or its full negation $\neg \Target{}$, or after reaching a maximum number of steps $N_\text{max}$.

\begin{wraptable}{r}{0.37\textwidth}
    \centering
    \caption{Connector types for each $\epsilon$.}
    \label{tab:action_edge_types}
    \begin{tabular}{|c|p{.1cm}p{.5cm}p{.05cm}p{.1cm}p{.1cm}|} \hline
         $\epsilon$& \multicolumn{5}{|c|}{Build $\wedge_{i}$  from  $(\epsilon, i, j)$} \\ \hline
         1& &  $\wedge_{i}$ & $\wedge$ & & $\wedge_{j}$\\
         2&  $\hspace{3pt}\neg$ & $\wedge_{i}$ & $\wedge$ &  & $\wedge_{j}$\\ 
         3& & $\wedge_{i}$ & $\wedge$ &$\hspace{3pt}\neg$   & $\wedge_{j}$\\
         4&  $\hspace{3pt}\neg$ & $\wedge_{i}$ & $\wedge$ & $\hspace{3pt}\neg$ &  $\wedge_{j}$\\ \hline
    \end{tabular}
\end{wraptable}
To effectively explore and prune the vast state space, our model comprises a policy and a value module  to assess intermediate states and strategically get closer to the desired target. 
As shown on Fig.~\ref{fig:Model}, our architecture consists of a shared core embedding the truth tables applying position encodings and $H$ transformer encoder layers. 
Then, the hidden embeddings are passed as input to the 4 stacked policy modules and to the value module.
The policy modules combined with a softmax produce a distribution over next actions, and the value module ending with $\operatorname{tanh}$ predicts an expected reward in $[-1, 1]$. 

\paragraph{Positional Encoding (PE)}
In natural language processing (NLP) tasks, PE allows transformers to capture a sequential relation in the inputs. 
In our setting, we do not need to hint the model about the graph structure graph, as the truth tables already convey all the necessary information.
Moreover, the self-attention operation should treat every node  equally, as there is no limitation regarding which two nodes can be combined to form a new node.
Nevertheless, the model should be able to treat differently the built nodes and the target truth table, leading us to introduce  two learnable
positional encoders to distinguish them (``Node PE'' on Fig.~\ref{fig:Model}).

\paragraph{Policy Module}
Transformers are the state-of-the-art architecture to handle sequential data.
These models particularly shine when trained to predict the next token in a sequence by outputting  a fixed-size tensor representing a sampling probability over a token glossary.
In our case though, the set of nodes that we can build grows at each step, as more pairs of nodes can be combined to produce the next node.
Inspired by NLP tasks, for which attention scores among related tokens are high, we use attention to guide next node generation. 
Thus, we directly  use the final self-attention map of four parallel policy modules to get the probability to connect any two nodes with specific edge types.
Each policy module has $P$ transformer encoder layers and outputs a final self-attention layer. 
In the last self-attention layer, we exclude the entry corresponding to the target node and returns the self-attention scores based on the existing AIG nodes embeddings. 
We aggregate the scores from the four policy modules and mask the ones associated to already built nodes and their negate versions. 
We finally apply a softmax to the remaining scores to produce a single probability distribution.


\paragraph{Value Module}
The value module is a critical component 
to assess how favorable a state is and therefore prevent expanding unpromising states.
Our value module consists of $S$ many transformer encoder layers.
As the quality of a state not only depends on the nodes that are present in the graph at that stage, but also on the target truth table that should be realized,  the value module also uses the two learnable type-based positional encoders introduced above.
After performing the embedding, we compute the cross-attention between the graph nodes and the target truth table, which yields a new vector representation of the target.
Finally, we feed this  vector to a linear layer producing a single value, which should reflect the quality of the current state.


\section{Training \OurMethod{}} \label{sec:training}
The sparse nature of the problem makes it practically impossible to discover functionally correct graphs when exploring the double exponential state space uniformly.
This challenge necessitates a more effective training approach for our \OurMethod{}.
To address this, we propose a two-stage training regimen consisting of a supervised pre-training stage to initialize the policy module, followed by an AlphaZero-style fine-tuning phase to improve the policy module and train the value module.
Although pre-training the policy module provides a good prior to predict the most useful next actions, simply following it does not guarantee that an AIG matching $\Target{}$ will be constructed due to the inherent difficulty of the problem.
This limitation highlights the importance of the fine-tuning stage, which aims to refine the policy module and leverage the value module for improved performance.


\subsection{Pre-Training} \label{sec:pre-training}

Just as large language models in NLP are pre-trained using next-token prediction, we can pre-train our transformer model on a next-node prediction task.
This approach requires a corpus of single-output AIGs that our model can learn to regenerate node by node, starting from the truth tables of the primary inputs and a target $\Target{}$.
Since the AIG generation problem has not yet been addressed by the ML community, there is no large, well-structured corpus of publicly available (truth table, AIG) pairs to perform this pre-training.
Therefore, we first curate a dataset of single-output AIGs for the pre-training of  \OurMethod{} by leveraging existing open-source collections of digital circuits.

\paragraph{Data Extraction}
To generate an AIG dataset with the desired input and output sizes, we utilize the EPFL benchmarks~\citep{EPFL-lsils}, which contain a collection of 20 real circuits realizing arithmetic and control functions.
On average, the arithmetic and control circuits have  175 inputs and 137 outputs with 22520 AND-nodes, as detailed on Table~\ref{tab:arithmetic_epfl_datasets} and ~\ref{tab:ranodmcontrol_epfl_datasets}.
Since these circuits have more inputs than the AIGs we aim to generate, we extract subgraphs, or \textit{cuts}, from them. 
A cut refers to a connected subset of nodes in the AIG that divides the graph into two disjoint parts. The root of a cut is the node to which all directed paths within the cut converge to and a \textit{leaf node} is a node in the cut that have at least one fanin outside of the cut.
By design, a cut forms a single-output AIG with a number of inputs corresponding to the number of leaves.
We defer to Appendix~\ref{sec:cut-extraction} the full description of the cut extraction method we develop to build a dataset of single-output AIGs.

\paragraph{Data Preparation}
Since our policy should predict the next action, i.e. the next node to add to a partial AIG, we need to convert the AIGs we load from our training dataset into a sequence of ground-truth actions.
As different series of actions can lead to the same graph with $N$ nodes, we first sort the nodes of the  training AIG we load into a topological order 
$\{I_1, \dots, I_n,\wedge_{n+1}, \dots, \wedge_N\}$
(or $\{v_1, \dots, v_N \}$ with node-type agnostic notation), where $\wedge_N$ is connected to the output $O$.
We also convert the AIG nodes into  truth tables, as described in section~\ref{sec:background}, and use the truth table of $O$ as the target  $\Target{}$. 
From the topological sequence of nodes, we build the sequence of actions that our policy should learn to perform when its goal is to generate $\Target{}$.
As mentioned in section~\ref{sec:method}, creating node $v_k = (\neg)v_{i} \ \wedge \ (\neg)v_{j}$, with $1 \leq i < j < k$,  corresponds to action $a_k = (\epsilon, i, j)$, whose first component $\epsilon \in \{1, 2, 3, 4\}$, indicates the types of the edges connecting $v_k$ to its parents, as detailed in Table~\ref{tab:action_edge_types}.
This procedure leaves us with the sequence of actions  $\mathcal{A} = \{a_{n+1}, a_{n+2}, ..., a_N\}$, starting with index $n+1$ since the $n$ primary inputs are given at the beginning of the AIG generation process. 

\begin{figure}[!t]
  \resizebox{\textwidth}{!}{\newcommand\bigzero{\makebox(0,0){\text{\huge0}}}
\newcommand*{\bord}{\multicolumn{1}{c|}{}}

\begin{tikzpicture}[scale=1.5, every node/.style={scale=1.2}]
    \tikzset{
        input/.style={circle, draw, minimum size=8mm, inner sep=0pt, line width=1.2pt},
        and/.style={circle, draw, fill=gray!30, minimum size=8mm, inner sep=0pt, line width=1.2pt},
        output/.style={circle, draw, fill=gray!60, minimum size=8mm, inner sep=0pt, line width=1.2pt},
        arrownode/.style={circle, draw, minimum size=6mm, inner sep=0pt},
        normal edge/.style={->, >=latex, line width=1.2pt},
        negated edge/.style={->, >=latex, dashed, line width=1.2pt},
        node distance=2cm and 2cm
    }

    \node[input] (I1) at (0, 0) {};
    \node[input, right=of I1, xshift=-1.5cm] (I2) {};
    \node[input, right=of I2, xshift=-1.5cm] (I3) {};
    \node[input, right=of I3, xshift=-1.5cm] (I4) {};
    \node[and, above=of $(I1.north)!0.5!(I2.north)$, yshift=-1cm] (A5) {};
    \node[and, above=of $(A5.north)!0.35!(I3.north)$, yshift=-1cm] (A6) {};
    \node[and, above=of $(I3.north)!0.5!(I4.north)$, yshift=-1cm] (A7) {};
    \node[and, above=of $(A7.north)!0.5!(A6.north)$, yshift=-1cm] (A8) {};
    \node[output, above=of A8, yshift=-.96cm] (O) {$O$};

    \draw[normal edge] (I1) -- (A5);
    \draw[negated edge] (I2) -- (A5);
    \draw[negated edge] (I3) -- (A7);
    \draw[normal edge] (I4) -- (A7);
    \draw[normal edge] (A5) -- (A6);
    \draw[normal edge] (I3) -- (A6);
    \draw[negated edge] (A7) -- (A8);
    \draw[normal edge] (A6) -- (A8);
    \draw[negated edge] (A8) -- (O);


    \node[input, right=of I4, xshift=-1.1cm] (I1wLabel) {$I_1$};
    \node[input, right=of I1wLabel, xshift=-1.5cm] (I2wLabel) {$I_2$};
    \node[input, right=of I2wLabel, xshift=-1.5cm] (I3wLabel) {$I_3$};
    \node[input, right=of I3wLabel, xshift=-1.5cm] (I4wLabel) {$I_4$};
    \node[and, above=of $(I1wLabel.north)!0.5!(I2wLabel.north)$, yshift=-1cm] (A5wLabel) {$\land_5$};
    \node[and, above=of $(I3wLabel.north)!0.5!(I4wLabel.north)$, yshift=-1cm] (A7wLabel) {$\land_7$};
    \node[and, above=of $(A5wLabel.north)!0.35!(I3wLabel.north)$, yshift=-1cm] (A6wLabel) {$\land_6$};
    \node[and, above=of $(A7wLabel.north)!0.5!(A6wLabel.north)$, yshift=-1cm] (A8wLabel) {$\land_8$};
    
    \node[output, above=of A8wLabel, yshift=-.96cm] (OwLabel) {$O$};

    \draw[normal edge, draw=color3] (I1wLabel) -- (A5wLabel);
    \draw[negated edge, draw=color3] (I2wLabel) -- (A5wLabel);
    \draw[negated edge, draw=color2] (I3wLabel) -- (A7wLabel);
    \draw[normal edge, draw=color2] (I4wLabel) -- (A7wLabel);
    \draw[normal edge, draw=color1] (A5wLabel) -- (A6wLabel);
    \draw[normal edge, draw=color1] (I3wLabel) -- (A6wLabel);
    \draw[negated edge, draw=color3] (A7wLabel) -- (A8wLabel);
    \draw[normal edge, draw=color3] (A6wLabel) -- (A8wLabel);
    \draw[negated edge] (A8wLabel) -- (OwLabel);

    \draw[->, >=latex, line width=2pt] ($(I4.north)+(0, 2)$)  -- ($(I1wLabel.north)+(0, 2)$) node[draw=none,midway,yshift=-.5cm] (node_ordering) {Node ordering};
    \node[draw=none, above=of node_ordering,yshift=-1.3cm] {Edge type marking};

    \node[draw=none, right=.7cm of A8wLabel, align=left, yshift=1.7cm] (actions) {$\mathcal{A} = \left\{ 
    \begin{array}{cccc}
        \textcolor{color3}{a_5 = (3, 1, 2)}, & \textcolor{color1}{a_6 = (1, 3, 5)}, &
         \textcolor{color2}{a_7 = (2, 3, 4)}, & \textcolor{color3}{a_8 = (3, 6, 7)}
    \end{array}
    \right\} $
    };

    \node[draw=none, below=.77cm of actions.west, anchor=west] (actions_matrix) {$\textbf{A} = (\textcolor{color1}{\boldsymbol{A}_1}, \textcolor{color2}{\boldsymbol{A}_2}, \textcolor{color3}{\boldsymbol{A}_3}, \textcolor{color4}{\boldsymbol{A}_4})$};

    \draw[black,thick,dotted, line width=1.2pt] ($(actions_matrix.center)+(0.32,0.27)$)  rectangle ($(actions_matrix.center)+(0.72,-0.27)$);
    
    \node[draw=none,below right=-.1cm of actions_matrix] (A3mat) {
    \setlength{\arraycolsep}{0pt} 
    $\begin{array}{ c }
    \textcolor{color3}{\left(\kern-\nulldelimiterspace
    \vphantom{\begin{array}{ c }
        \bord  \\[3pt]
        \bord \\[3pt]
        \bord \\[3pt]
        \bord \\[3pt]
        \bord \\[3pt]
        \bord \\[3pt]
        \bord \\[4pt]
                0 \\[2pt]
    \end{array}}
    \right.}
        \\[13pt]
        \\[13pt]
      \end{array}
    \setlength{\arraycolsep}{5pt}
    \arrayrulecolor{color3}
    \begin{array}{ccccccc} \cline{1-1}
         \bord &  \textcolor{color3}{1} & \textcolor{color3}{0} & \textcolor{color3}{0} & \textcolor{color3}{0} & \textcolor{color3}{0} & \textcolor{color3}{0} \\[3pt] \cline{2-2} 
          & \bord & \textcolor{color3}{0} & \textcolor{color3}{0} & \textcolor{color3}{0} & \textcolor{color3}{0} & \textcolor{color3}{0} \\[3pt] \cline{3-3} 
          &  & \bord & \textcolor{color3}{0} & \textcolor{color3}{0} & \textcolor{color3}{0} & \textcolor{color3}{0} \\[3pt] \cline{4-4} 
          &  &  & \bord & \textcolor{color3}{0} & \textcolor{color3}{0} & \textcolor{color3}{0} \\[3pt] \cline{5-5} 
          &  &  \textcolor{color3}{\bigzero} &  & \bord & \textcolor{color3}{0} & \textcolor{color3}{0} \\[3pt] \cline{6-6} 
          &  &  &  &   & \bord &  \textcolor{color3}{1} \\[3pt] \cline{7-7} 
          & &  &  &  &  & \bord \\[3pt]
           & &  &  &  &  & \\[1pt]
         I_1 & I_2 & I_3 & I_4 & \wedge_5  & \wedge_6  & \wedge_7 \\[3pt]
       \end{array}
       \arrayrulecolor{black}
       \setlength{\arraycolsep}{0pt} 
    \ \begin{array}{ c }
    \textcolor{color3}{\left. \kern-\nulldelimiterspace
    \vphantom{\begin{array}{ c }
        0  \\[3pt]
        0 \\[3pt]
        0 \\[3pt]
        0 \\[3pt]
        0 \\[3pt]
        0 \\[3pt]
        0 \\[4pt]
        0 \\[2pt]
    \end{array}}
    \right)}
        \\[13pt]
        \\[13pt]
      \end{array}
    \setlength{\arraycolsep}{5pt}
       \begin{array}{c}
        I_1 \\[3pt]
        I_2 \\[3pt]
        I_3 \\[3pt]
        I_4 \\[3pt]
        \wedge_5 \\[3pt]
        \wedge_6 \\[3pt]
        \wedge_7 \\[3pt]
       \\[1pt]
        \\[3pt]
       \end{array}
       $
       };

       \draw[->, >=latex, line width=1pt] let \p1 = (A3mat.west), \p2 = (actions_matrix.center)
           in ($(actions_matrix.center)+(0.52,-0.27)$) -- ($(actions_matrix.center)+(0.52,\y1) - (0.,\y2)$) -- ($(\x1, \y1)$);

     \draw[black,thick,dotted, line width=1.8pt] ($(A3mat.center)+(-2.32,1.8)$)  rectangle ($(A3mat.center)+(-1.22,.9)$) node[yshift=1.4cm, xshift=.6cm, color=black] {Submatrix: $\boldsymbol{A}_{3, 1:2, 1:2}$};
    
\end{tikzpicture}}
  \caption{We start data pre-processing by sorting the AIG nodes in topological order.
  Then, we identify the action types $\epsilon \in \{1, 2, 3, 4\}$ based on the edges. 
  Next, we build the sequence of actions $\mathcal{A}$ and generate the global action tensor $\boldsymbol{A} = (\boldsymbol{A}_1, \boldsymbol{A}_2, \boldsymbol{A}_3,\boldsymbol{A}_4)$.
  We highlight the structure of $\boldsymbol{A}_3$, which contains a $1$ at entries $(1, 2)$ and $(6, 7)$ for the generation of $\wedge_5$ and $\wedge_8$ 
  (actions $a_5$ and $a_8$).
  }
  \label{fig:actions}
\end{figure}
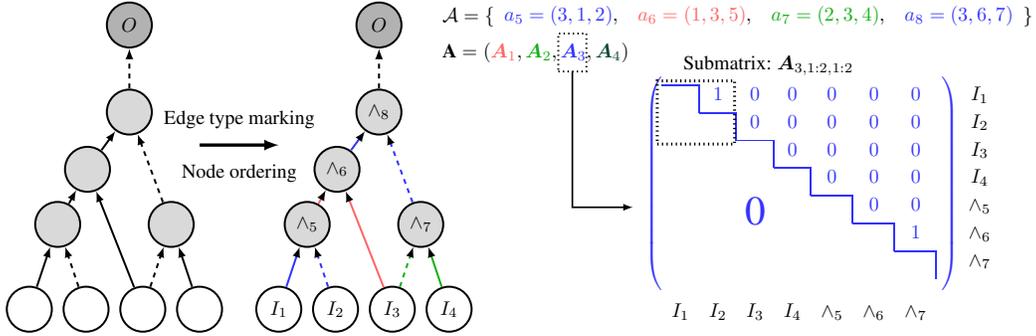

To efficiently generate target action distributions, we aggregate all the actions into a sparse $3$-dimensional tensor $\mathbf{A} = \left(\mathbf{A}_1, \mathbf{A}_2, \mathbf{A}_3, \mathbf{A}_4\right) $ where each element $\mathbf{A}_\epsilon$ is a $N \times N$ matrix representing the actions with connection type $\epsilon$.
The value of the entry $(i, j)$ of $\mathbf{A}_\epsilon$ is set to $1$ if $(\epsilon, i, j)$ belongs to $\mathcal{A}$ and to zero otherwise.
Thus, if all the nodes up to $v_{k}$ are already built, considering each submatrix $\mathbf{A}_{\epsilon, 1:k, 1:k}$ taking the first $k$ rows and $k$ columns of $\mathbf{A}_\epsilon$ allows to easily identify which nodes with connection $\epsilon$ we could build next.
Taking the submatrices for all values of  $\epsilon$, we obtain the target action distribution by setting the entries corresponding to already performed actions at 0, and normalizing the resulting tensor.
Fig.~\ref{fig:actions} illustrates this action tensor building procedure.

\paragraph{Data Augmentation}
The first data augmentation we employ consists of using the same AIG for both  targets $\Target{}$ and $\neg \Target{}$.
This is valid because  we can generate one target or the other by connecting the final node $\wedge_N$ with the output $O$ using a regular or an inverter edge.
Our second data augmentation leverages the fact that any order of truth table rows is valid, provided that the same order is used for all the nodes.
Since our \OurMethod{}'s inputs are truth tables, it is desirable for the model to be invariant to row permutations.
Formally, the model should generate the same next-action prediction whether it receives the truth tables $T_1, \dots, T_N$ and  $\Target{}$, where $T_i = (t^{(1)}_i, \dots, t^{(2^n)}_i)\in\{0, 1\}^{2^n}$, or when it gets the permuted truth tables $\sigma(T_1), \dots, \sigma(T_N), \sigma(\Target{})$  where $\sigma$ is a permutation in $\mathbb{S}_{2^n}$ and $\sigma(T_i) = \left(t_i^{(\sigma(1))}, \dots, t_i^{(\sigma(2^n))}\right)$.
As structurally encoding this invariance into our policy architecture would be computationally too expensive, we apply random permutations to the inputs of our
model during the training, which does not impact the other metadata introduced in the previous section.


\paragraph{Pre-Training Flow}
With the prepared augmented data, we can proceed to train our policy module to match the ground-truth next-action distributions of our training set. 
For training loss, we experimented with KL divergence and cross-entropy, both of which measure the distance between two probability distributions. 
In practice, KL divergence loss yielded better results.
Besides, the backbone of our model being a  transformer, we implement a custom masking strategy during training. to maintain causality in the auto-regressive generation process.
Since the primary inputs and the target truth table are available from the start, and as there is no causality for their existence, 
we allow full attention for their embeddings, and only apply a causal mask for the rest of the nodes.


\subsection{Fine-Tuning}

Fine-tuning aims to align the value and policy module so they operate effectively together.
Unlike the policy module, we cannot  properly initialize the value module during pre-training as the generated dataset only contains successful examples, which would mislead the value module to consider that all states are ``good''.
Skipping pre-training, though, would lead to a random exploration of the vast search space of truth tables, which would likely result in encountering only ``bad'' states, preventing the model from learning what a ``good'' state is.
Therefore, the most viable option to train our value module is through experience, by performing searches with a pre-trained policy module. 
We utilize AlphaZero as the orchestration framework to refine the policy and value modules.

\paragraph{AlphaZero}
AlphaZero has demonstrated remarkable success in board games with enormous state spaces, such as chess ($10^{44}$) and Go ($10^{170}$).
Since truth tables features similar state space problem, we adapt AlphaZero's effective search and pruning capabilities to navigate AIG generation.
By combining a policy module to propose actions and a value module to evaluate state viability, AlphaZero  strikes a balance between exploitation and exploration.
We adapt and modify the selection strategy, predictor upper confidence bound applied to trees (PUCT) used by AlphaZero, as follows:
\begin{equation*}
    \operatorname{PUCT}(s,a) = Q(s,a) + b\ Q(s,a;\theta) + c\ P(s,a; \theta)
    \frac{\sqrt{\sum_{a}N(s,a)}}{N(s,a) + 1}
\end{equation*}
where, $Q(s,a)$ represents the propagated discounted discovered reward, while $Q(s,a; \theta)$ represents the predicted expected Q-value, $P(s,a; \theta)$ is the policy module's probability distribution, $N(s,a)$ tracks state visitations, and $b$ and $c$ are parameters balancing exploration and exploitation.
Computing $Q(s_t,a;\theta)$ for every action is too expensive, so we initialize $Q(s_t,a)=Q(s_t,a;\theta)=0$, perform the action that maximizes $\operatorname{PUCT}(s_t,a)$, and only compute the value of the state $Q(s_{t+1})$ once we visit it.
The term $Q(s,a) + bQ(s,a;\theta)$ represents the exploitation in PUCT, as if during search our method discovers a "good" state or a terminal state, we exploit it and focus the search locally to discover more compact designs.
The term $P(s,a;\theta)$ suggests actions to perform, but the term $\nicefrac{\sqrt{\sum_{a}N(s,a)}}{(N(s,a) + 1)}$ promotes exploration.

AlphaZero stores intermediate results and metadata, such as $Q(s,a)$, $Q(s,a;\theta)$, $P(s,a;\theta)$, and $N(s,a)$, in the nodes visited during MCTS.
These nodes are associated with states and form a tree, where edges indicate the actions performed to reach each node-state pair.
When simulation starts, we mark the initial state as the root node, compute the action distribution, and inject Dirichlet noise.
During simulation, AlphaZero follows PUCT to choose actions and continues until meeting one of the three following stopping conditions: encountering a state $s$ that is not expanded, reaching a maximum number of steps, or arriving at a terminal state.
If the state is not expanded, we need to compute $Q(s;\theta)$ and $P(s,a;\theta)$ for that state and we back-propagate $Q(s;\theta)$ to the previous MCTS nodes, and increment $N(s,a)$.
Once we complete the given number of simulations, AlphaZero applies the most visited action, ${\operatorname{argmax}}_{a\in \mathcal{A}}\ N(s,a)$.
In our case, we rather follow the observed discounted reward ${\operatorname{argmax}}_{a\in \mathcal{A}}\ Q(s,a)$ as we find the visitation count signal too noisy given our simulation budget. 


\paragraph{Fine-Tuning Flow}
Generating trajectories for millions of truth tables is computationally challenging.
Thus, to best exploit our resources, our fine-tuning regimen consists of data collection and model training processes.
These data collectors generate trajectories and add their findings to a fixed length replay buffer.
Under the hood, the data collectors store the metadata, including truth tables and discovered reward $Q(s)$, of the MCTS root node for each step in the trajectory in the replay buffer.
Successful trajectories receive a reward of $1$, while failed ones receive $-\min{(h_d(T_N, \Target{}), h_d(T_N, \neg \Target{}))}$, where $h_d$ is the Hamming distance and $T_N$ is the last generated truth table.

The trainer process randomly samples data from the replay buffer and uses the truth tables as input for the model.
Since the value module aims to predict the Q-value of a state, the goal is to minimize the mean squared error (MSE) between the predicted value and the retrieved $Q(s)$.
The target distribution for the policy module is the normalized number of visitations $\nicefrac{N(s,a)}{\sum_a{N(s,a)}}$.
Similar to pre-training, we minimize the KL-divergence between the output of the policy module and the target probability distribution.
Finally, the trainer process broadcasts the updated 
weights asynchronously to the data collectors after a given number of training steps.


\section{Experimental Evaluation}\label{sec:exp_evaluation}

We introduce the implementation details such as model and search hyperparameters, datasets and baseline methods in section \ref{sec:exp_settings} and \ref{sec:exp_baselines}. 
We compare the effectiveness of \OurMethod{} against several baselines in \ref{sec:exp_quality}. Finally, section \ref{sec:exp_ablation} presents a study on the impact of number of simulations.

\subsection{Experimental Setup}\label{sec:exp_settings}
We train and evaluate \OurMethod{} on the $8$-input truth tables that we randomly  extract from the EPFL benchmark, as described in section~\ref{sec:pre-training}. 
We specifically choose to test on these circuits, since they correspond to real-world Boolean functions that have more practical interest than uniformly random truth-tables.
In total, we extract $1.8$ million AIGs with an average number of AND-nodes of $10.08$.
We pre-train \OurMethod{} with a batch size of 1024 for 250 epochs, and finetune the model until it converges.
Our model architecture is as depicted on Fig.~\ref{fig:Model} and uses transformer blocks following Llama 3~\citep{meta2024llama3} structure. 
We use are $H=4$ and $P=S=3$ transformer blocks for the different parts with $16$ heads and an intermediate embedding size of $4096$, summing to $51.6$ million parameters. 
During pre-training, we apply a uniformly sampled permutation transform to the training sequences and a target negation transform both with probability $50\%$. 

\subsection{Baselines}\label{sec:exp_baselines}

We derive each truth table in our test from the primary output of an extracted cut.
Therefore, we can use that cut as a baseline, denoted as \texttt{Cut}, as it is an AIG realizing the target truth table.
Moreover, 
we leverage a popular logic optimization flow, \texttt{resyn2}, that applies multiple operators to optimize the \texttt{Cut} AIGs.
While \texttt{resyn2} cannot guarantee the optimality of the resulting AIGs, it provides a reliable optimality proxy for the current graph sizes, which we represent as a horizontal line with the tick label \textbf{O}, when applicable.
Additionally, we compare \OurMethod{} against the state-of-the-art open-source logic synthesis tool \texttt{ABC}.
This library applies a series of Boolean algebra transformations
to generate an AIG from a truth table. 
The sequence of commands we use
in \texttt{ABC} is as follows: \texttt{read\_truth -x [truth table]; collapse; sop; strash; write [outfile]}.
We also apply \texttt{resyn2} to the obtained AIGs, which we denote as \texttt{ABC+resyn2}.
Finally, we compare \OurMethod{} against Boolformer with default beam size of 10. 
This  learned method  produces an optimized Boolean expression given a truth table, and we convert this expression into an AIG without introducing any additional logic redundancy.


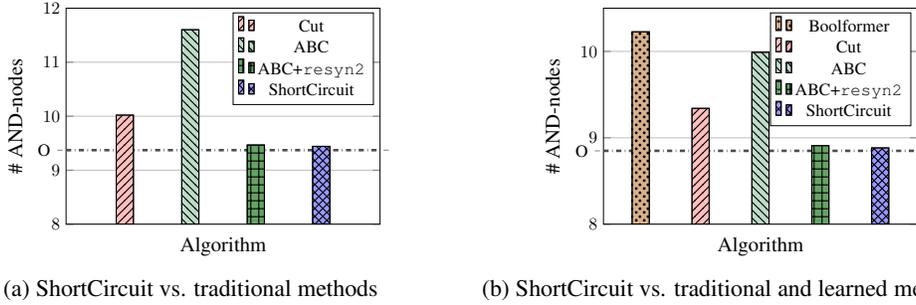
\begin{figure}[ht]
\centering
    \begin{subfigure}[b]{0.48\textwidth}
        \centering
        \resizebox{0.8\textwidth}{!}{\definecolor{forestgreen}{RGB}{34,139,34}
\definecolor{darkgreen}{RGB}{0,100,0}
\definecolor{huntergreen}{RGB}{34,139,69}

\begin{tikzpicture}
\begin{axis}[
    height=6cm,
    width=8cm,
    ybar, 
    xlabel={ \large{Algorithm}}, 
    ylabel={\large{\# AND-nodes}}, 
    ymin=8, 
    ymax=12, 
    xtick=\empty, 
    xmin=0,
    xmax=7,
    grid=both,
    xmajorgrids=false,
    ymajorgrids=true,
    extra y ticks={\QOptimal},
    extra y tick labels={O},
    extra y tick style={
        dashdotted,
        tick align=outside,
        major tick length=5pt,
        line width=1.5pt,
        grid style={draw=black!70, line width=1.5pt} 
    },
]

\addplot[draw=black, fill=red!25, postaction={pattern=north east lines, pattern color=black}, error bars/.cd, y dir=both, y explicit, error bar style={black}] coordinates {(2, \QCut)};
\addplot[draw=black, fill=huntergreen!30, postaction={pattern=north west lines, pattern color=black}, error bars/.cd, y dir=both, y explicit, error bar style={black}] coordinates {(3, \QABC)};
\addplot[draw=black, fill=darkgreen!60, postaction={pattern=grid, pattern color=black}, error bars/.cd, y dir=both, y explicit, error bar style={black}] coordinates {(4, \QABCOpt)};
\addplot[draw=black, fill=blue!40, postaction={pattern=crosshatch, pattern color=black}, error bars/.cd, y dir=both, y explicit, error bar style={black}] coordinates {(5, \QOurMethod)};

\legend{Cut, ABC, ABC+\texttt{resyn2}, \OurMethod{}} 
\end{axis}
\end{tikzpicture}}
        \caption{\OurMethod{} vs. traditional methods\label{fig:aig_size}}
    \end{subfigure}
\quad
    \begin{subfigure}[b]{0.48\textwidth}
        \centering
        \resizebox{0.8\textwidth}{!}{\definecolor{forestgreen}{RGB}{34,139,34}
\definecolor{darkgreen}{RGB}{0,100,0}
\definecolor{huntergreen}{RGB}{34,139,69}

\begin{tikzpicture}
\begin{axis}[
    ybar, 
    height=6cm,
    width=8cm,
    xlabel={\large{Algorithm}}, 
    ylabel={\large{\# AND-nodes}}, 
    ymin=8, 
    ymax=10.5, 
    xtick=\empty, 
    xmin=0,
    xmax=8,
    grid=both,
    xmajorgrids=false,
    ymajorgrids=true,
    extra y ticks={\QOptimalB},
    extra y tick labels={O},
    extra y tick style={
        dashdotted,
        tick align=outside,
        major tick length=5pt,
        line width=1.5pt,
        grid style={draw=black!70, line width=1.5pt} 
    },
]

\addplot[draw=black, fill=brown!60, postaction={pattern=crosshatch dots, pattern color=black}, error bars/.cd, y dir=both, y explicit, error bar style={black}] coordinates {(2, \QBoolformer)};
\addplot[draw=black, fill=red!25, postaction={pattern=north east lines, pattern color=black}, error bars/.cd, y dir=both, y explicit, error bar style={black}] coordinates {(3, \QCutB)};
\addplot[draw=black, fill=huntergreen!30, postaction={pattern=north west lines, pattern color=black}, error bars/.cd, y dir=both, y explicit, error bar style={black}] coordinates {(4, \QABCB)};
\addplot[draw=black, fill=darkgreen!60, postaction={pattern=grid, pattern color=black}, error bars/.cd, y dir=both, y explicit, error bar style={black}] coordinates {(5, \QABCOptB)};
\addplot[draw=black, fill=blue!40, postaction={pattern=crosshatch, pattern color=black}, error bars/.cd, y dir=both, y explicit, error bar style={black}] coordinates {(6, \QOurMethodB)};

\legend{Boolformer, Cut, ABC, ABC+\texttt{resyn2}, \OurMethod{}} 
\end{axis}
\end{tikzpicture}}
        \caption{\OurMethod{} vs. traditional and learned methods}
        \label{fig:aig_size_bool}
    \end{subfigure}
\caption{Average number of AND-nodes for the successfully generated AIGs across several baselines.}
\end{figure}

\subsection{Generation Quality Experiments} \label{sec:exp_quality}

We evaluate \OurMethod{} against our baselines on $500$ truth tables associated with randomly sampled AIGs from the EPFL benchmarks. 
We allow \OurMethod{} to attempt to generate a circuit with up to $30$ AND-nodes.
\OurMethod{} performs $8$ MCTS simulations and generates up to $20$ AND-nodes in each simulation, before performing an action.
The success rate of \OurMethod{} on this test set is $\QSuccessRate \%$.

Fig.~\ref{fig:aig_size} compares the average size of the successfully generated circuits compared to \texttt{Cut}, \texttt{ABC}, and \texttt{ABC+resyn2}.
\OurMethod{} generates compact AIGs, near-optimal as indicated by the horizontal line at the tick \textbf{O}, with an average time per generation of $1.83$s.
Specifically, \OurMethod{} produces circuits with $\fpeval{round(\QOurMethod,2)}$ AND-nodes on average, which is $\percentdiff{\QOptimal}{\QOurMethod}\%$ larger than the optimal ones that have $\fpeval{round(\QOptimal,2)}$ AND-nodes on average.
Furthermore, the AIGs generated by \OurMethod{} are significantly smaller than the ones from the baselines \texttt{Cut} and \texttt{ABC}, achieving a size reduction of $\percentdiff{\QCut}{\QOurMethod}\%$ and $\percentdiff{\QABC}{\QOurMethod}\%$, respectively.
Finally, \OurMethod{} produces slightly smaller AIGs than \texttt{ABC+resyn2} by $\percentdiff{\QABCOpt}{\QOurMethod}\%$. 
While these results demonstrate the effectiveness of \OurMethod{}, changes of parameters, such as increasing the AND-node limits, can further improve the success rate, albeit causing extra runtime.

Fig. \ref{fig:aig_size_bool} compares the AIG sizes against the same baselines as above with the addition of Boolformer.
Boolformer successfully generated $85\%$ of the given truth tables, so we report the results only on the successfully generated by both learned methods.
\OurMethod{} still maintains similarly good performance on this test subset as on the previous set shown on Fig.~\ref{fig:aig_size}.
Boolformer produces larger AIGs and fails in particular to generate the larger AIGs, but it produces a Boolean expression in about $0.75s$.
\OurMethod{}'s AIGs are $\percentdiff{\QBoolformer}{\QOurMethodB}\%$ smaller while the running time, $\QRuntime$s, is comparable.

\subsection{Impact of Number of Simulations}\label{sec:exp_ablation}

\begin{figure}[ht]
\begin{floatrow}
\ffigbox{%
  \resizebox{0.3\textwidth}{!}{\begin{tikzpicture}
    \begin{semilogxaxis}[
        xlabel=\Large{\# MCTS Simulations},
        ylabel=\Large{Success$(\%)$},
        xmin=0.7, xmax=300,
        ymin=90, ymax=100,
        grid=both,
        minor grid style={gray!25},
        major grid style={gray!50},
        log basis x=2,
        error bars/y dir=both,
        error bars/y explicit,
        point meta=none,
    ]
    \addplot[scatter,only marks,mark=square*,mark options={black}, draw=black]
    table[x=x,y=y,y error plus=plus,y error minus=minus] {
        x  y  plus minus
        1 92.2 2.3505844062659764 2.3505844062659764
        2 97.2 1.44602312058979 1.44602312058979
        4 97.4 1.3948555708406363 1.3948555708406363
        8 98.0 1.227131556807215 1.227131556807215
        16 97.8 1.285712471636291 1.285712471636291
        32 98.2 1.1653465245547334 1.1653465245547334
        64 98.4 1.099817509162393 1.099817509162393
        128 98.6 1.0298300593870935 1.0298300593870935
        256 98.6 1.0298300593870935 1.0298300593870935
    };
    \end{semilogxaxis}
\end{tikzpicture}}
}{%
  \caption{\OurMethod{}'s success rate vs. number of MCTS simulations per action.}%
  \label{fig:ablation_mcts_success}
}
\ffigbox{%
    \resizebox{0.48\textwidth}{!}{\begin{tikzpicture}
    \begin{semilogxaxis}[
        xlabel=\Large{Average runtime per generation (s)},
        ylabel=\Large{\# AND-nodes},
        xmin=0.1, xmax=100,
        ymin=9, ymax=10,
        grid=both,
        minor grid style={gray!25},
        major grid style={gray!50},
        log basis x=10,
        width=0.9\textwidth,
        height=0.45\textwidth,
    ]
\addplot[scatter, only marks, mark=x, mark size=4pt, nodes near coords*={\pgfplotspointmeta}, 
every node near coord/.append style={yshift=4pt,font=\normalsize},
point meta=explicit symbolic]
    table[
        x=x,
        y=y,
        meta=label
    ] {
        x  y  x_err y_err label
        0.11 9.633 0.07 2.441 1
        0.99 9.490 0.97 2.445 2
        1.93 9.458 1.87 2.333 4
        3.50 9.442 3.49 2.272 8
        6.39 9.438 7.52 2.312 16
        11.85 9.409 13.16 2.111 32
        21.66 9.430 26.15 2.081 64
        44.77 9.411 49.52 1.969 128
        71.86 9.375 82.52 1.845 256
    };
    \end{semilogxaxis}
\end{tikzpicture}}
}{
    \caption{Average \#AND-nodes per AIG vs. generation time for \OurMethod{} with varying MCTS simulations marked on top of each data point.}%
    \label{fig:ablation_mcts_quality}
}
    
\end{floatrow}
\end{figure}

To better understand the role of MCTS simulations in the performance of \OurMethod{}, we investigate their impact on the success rate, the circuit size, and the execution time.
Using the same test set of $500$ truth tables as in section~\ref{sec:exp_quality}, we evaluate our method when it solely relies on its policy module.
Fig.~\ref{fig:ablation_mcts_success} illustrates how the success rate evolves as the number of MCTS simulations increases from 1 to 256.
For clarity, we append an integer $i$ next to our method's name 
(\OurMethod{}$[i]$) 
to indicate the number of simulations we perform.

When using only 1 simulation, \OurMethod{}[1] performs a greedy search, where the policy selects the most likely action, $a = \operatorname{argmax}_{a\in\mathcal{A}} P(s,a;\theta)$.
This greedy strategy yields a low success rate of $\pgfkeysvalueof{/Ablation/S1}\%$ but benefits from a very short generation time of just $\pgfkeysvalueof{/Ablation/T1}s$.
On the other end, \OurMethod{}$[256]$ achieves a significantly higher success rate of $\pgfkeysvalueof{/Ablation/S256}\%$,  albeit with a much longer running time of $\pgfkeysvalueof{/Ablation/T256}s$.
Increasing the number of simulations enables the model to explore a larger -- but still limited -- portion of the solution space, resulting in higher success rates and the discovery of more compact designs.
For example, \OurMethod{}$[256]$ generated as ``optimal'' AIGs as \textbf{O}.
Fig.~\ref{fig:ablation_mcts_quality} highlights this trade-off by revealing a Pareto front, suggesting that we can adjust the number of MCTS simulations to achieve the desired balance between success rate, design quality, and running time.



\section{Conclusion}
In this paper, we introduced \OurMethod{}, a novel transformer-based architecture for generating AIGs from a target truth table.
Our approach combines a structurally aware transformer model with an AlphaZero-inspired policy variant, enabling efficient navigation through the doubly exponential state space associated with truth tables.
In our experiments, we demonstrated the effectiveness of \OurMethod{} in producing high-quality AIGs that are significantly smaller than those generated by one of the state-of-the-art logic synthesis tools \texttt{ABC} and the trained Boolformer.
Specifically, our method achieved a relative size reduction of $\percentdiff{\QABC}{\QOurMethod}\%$, and $\percentdiff{\QOurMethodB}{\QBoolformer}\%$ respectively.

This work contributes to the expanding field of ML applications in chip design, showcasing the potential of deep learning to revitalize the field with new perspectives.
We demonstrated that it is possible to generate high-quality AIGs from truth tables, paving the way for future research in this area.
Future work will focus on extending \OurMethod{} to handle AIGs with multiple outputs, integrating our approach with existing logic synthesis tools, and exploring its application in industrial settings.
Finally, our goal is to enable the creation of more efficient, scalable, and innovative computing systems, and we believe that  \OurMethod{} is an important step towards realizing this vision.

\bibliographystyle{unsrtnat}
\bibliography{ref}

\begin{thebibliography}{45}
\providecommand{\natexlab}[1]{#1}
\providecommand{\url}[1]{\texttt{#1}}
\expandafter\ifx\csname urlstyle\endcsname\relax
  \providecommand{\doi}[1]{doi: #1}\else
  \providecommand{\doi}{doi: \begingroup \urlstyle{rm}\Url}\fi

\bibitem[Mishchenko and Brayton(2006)]{Mishchenko2006ScalableLS}
Alan Mishchenko and Robert~K. Brayton.
\newblock Scalable logic synthesis using a simple circuit structure.
\newblock 2006.
\newblock URL \url{https://api.semanticscholar.org/CorpusID:8597391}.

\bibitem[Wolf et~al.(2013)Wolf, Glaser, and Kepler]{Wolf2013YosysAFV}
Clifford Wolf, Johann Glaser, and Johannes Kepler.
\newblock Yosys-a free verilog synthesis suite.
\newblock 2013.
\newblock URL \url{https://api.semanticscholar.org/CorpusID:202611483}.

\bibitem[Huang et~al.(2021)Huang, Hu, He, Liu, Ma, Shen, Wu, Xu, Zhang, Zhong,
  Ning, Ma, Yang, Yu, Yang, and Wang]{Huang2021SurveyML4EDA}
Guyue Huang, Jingbo Hu, Yifan He, Jialong Liu, Mingyuan Ma, Zhaoyang Shen,
  Juejian Wu, Yuanfan Xu, Hengrui Zhang, Kai Zhong, Xuefei Ning, Yuzhe Ma,
  Haoyu Yang, Bei Yu, Huazhong Yang, and Yu~Wang.
\newblock Machine learning for electronic design automation: A survey.
\newblock \emph{ACM Trans. Design Autom. Electr. Syst.}, 26:\penalty0
  40:1--40:46, 2021.
\newblock URL \url{https://api.semanticscholar.org/CorpusID:231839647}.

\bibitem[Gubbi et~al.(2022)Gubbi, Beheshti-Shirazi, Sheaves, Salehi, PD,
  Rafatirad, Sasan, and Homayoun]{Gubbi2022SurveyML4EDA}
Kevin~Immanuel Gubbi, Sayed~Aresh Beheshti-Shirazi, Tyler Sheaves, Soheil
  Salehi, Sai~Manoj PD, Setareh Rafatirad, Avesta Sasan, and Houman Homayoun.
\newblock Survey of machine learning for electronic design automation.
\newblock In \emph{Proceedings of the Great Lakes Symposium on VLSI 2022},
  GLSVLSI '22, page 513–518, New York, NY, USA, 2022. Association for
  Computing Machinery.
\newblock ISBN 9781450393225.
\newblock \doi{10.1145/3526241.3530834}.
\newblock URL \url{https://doi.org/10.1145/3526241.3530834}.

\bibitem[Ward et~al.(2012)Ward, Kim, Viswanathan, Li, Alpert, Swartzlander, and
  Pan]{Ward2012KeepIS}
Samuel~I. Ward, Myung-Chul Kim, Natarajan Viswanathan, Zhuo Li, Charles~J.
  Alpert, Earl~E. Swartzlander, and David~Z. Pan.
\newblock Keep it straight: teaching placement how to better handle designs
  with datapaths.
\newblock In \emph{ACM International Symposium on Physical Design}, 2012.
\newblock URL \url{https://api.semanticscholar.org/CorpusID:8570378}.

\bibitem[Alawieh et~al.(2020)Alawieh, Li, Lin, Singhal, Iyer, and
  Pan]{Alawieh2020Routing}
Mohamed~Baker Alawieh, Wuxi Li, Yibo Lin, Love Singhal, Mahesh~A. Iyer, and
  David~Z. Pan.
\newblock High-definition routing congestion prediction for large-scale fpgas.
\newblock In \emph{2020 25th Asia and South Pacific Design Automation
  Conference (ASP-DAC)}, pages 26--31, 2020.
\newblock \doi{10.1109/ASP-DAC47756.2020.9045178}.

\bibitem[Tu et~al.(2024)Tu, Tang, Yu, Josipovi{\'{c}}, and
  Chu]{Tu2024AI4LogicSynthesis}
Kaihui Tu, Xifan Tang, Cunxi Yu, Lana Josipovi{\'{c}}, and Zhufei Chu.
\newblock \emph{Logic Synthesis}, pages 135--164.
\newblock Springer Nature Singapore, Singapore, 2024.
\newblock ISBN 978-981-99-7755-0.
\newblock \doi{10.1007/978-981-99-7755-0_9}.
\newblock URL \url{https://doi.org/10.1007/978-981-99-7755-0_9}.

\bibitem[d'Ascoli et~al.(2024)d'Ascoli, Bengio, Susskind, and
  Abbe]{ascoli2024boolformer}
St{\'e}phane d'Ascoli, Samy Bengio, Joshua~M. Susskind, and Emmanuel Abbe.
\newblock Boolformer: Symbolic regression of logic functions with transformers,
  2024.
\newblock URL \url{https://openreview.net/forum?id=wmzFZ9lJrD}.

\bibitem[Li et~al.(2024{\natexlab{a}})Li, Li, Chen, Zhang, Yuan, and
  Wang]{Li2024CircuitTransformer}
Xihan Li, Xing Li, Lei Chen, Xing Zhang, Mingxuan Yuan, and Jun Wang.
\newblock Circuit transformer: End-to-end circuit design by predicting the next
  gate.
\newblock \emph{ArXiv}, abs/2403.13838, 2024{\natexlab{a}}.
\newblock URL \url{https://api.semanticscholar.org/CorpusID:268553512}.

\bibitem[Dong et~al.(2023)Dong, Cao, Zhang, Tao, Chen, and
  Zhang]{dong2023cktgnn}
Zehao Dong, Weidong Cao, Muhan Zhang, Dacheng Tao, Yixin Chen, and Xuan Zhang.
\newblock Cktgnn: Circuit graph neural network for electronic design
  automation.
\newblock \emph{arXiv preprint arXiv:2308.16406}, 2023.

\bibitem[Silver et~al.(2016)Silver, Huang, Maddison, Guez, Sifre, van~den
  Driessche, Schrittwieser, Antonoglou, Panneershelvam, Lanctot, Dieleman,
  Grewe, Nham, Kalchbrenner, Sutskever, Lillicrap, Leach, Kavukcuoglu, Graepel,
  and Hassabis]{Silver2016AlphaGO}
David Silver, Aja Huang, Chris~J. Maddison, Arthur Guez, Laurent Sifre, George
  van~den Driessche, Julian Schrittwieser, Ioannis Antonoglou, Vedavyas
  Panneershelvam, Marc Lanctot, Sander Dieleman, Dominik Grewe, John Nham, Nal
  Kalchbrenner, Ilya Sutskever, Timothy~P. Lillicrap, Madeleine Leach, Koray
  Kavukcuoglu, Thore Graepel, and Demis Hassabis.
\newblock Mastering the game of go with deep neural networks and tree search.
\newblock \emph{Nat.}, 529\penalty0 (7587):\penalty0 484--489, 2016.
\newblock \doi{10.1038/NATURE16961}.
\newblock URL \url{https://doi.org/10.1038/nature16961}.

\bibitem[Silver et~al.(2017)Silver, Hubert, Schrittwieser, Antonoglou, Lai,
  Guez, Lanctot, Sifre, Kumaran, Graepel, Lillicrap, Simonyan, and
  Hassabis]{Silver2017AlphaZero}
David Silver, Thomas Hubert, Julian Schrittwieser, Ioannis Antonoglou, Matthew
  Lai, Arthur Guez, Marc Lanctot, Laurent Sifre, Dharshan Kumaran, Thore
  Graepel, Timothy~P. Lillicrap, Karen Simonyan, and Demis Hassabis.
\newblock Mastering chess and shogi by self-play with a general reinforcement
  learning algorithm.
\newblock \emph{CoRR}, abs/1712.01815, 2017.
\newblock URL \url{http://arxiv.org/abs/1712.01815}.

\bibitem[Jumper et~al.(2021)Jumper, Evans, Pritzel, Green, Figurnov,
  Ronneberger, Tunyasuvunakool, Bates, Ž{\'i}dek, Potapenko, Bridgland, Meyer,
  Kohl, Ballard, Cowie, Romera-Paredes, Nikolov, Jain, Adler, Back, Petersen,
  Reiman, Clancy, Zielinski, Steinegger, Pacholska, Berghammer, Bodenstein,
  Silver, Vinyals, Senior, Kavukcuoglu, Kohli, and
  Hassabis]{Jumper2021AlphaFold}
John~M. Jumper, Richard Evans, Alexander Pritzel, Tim Green, Michael Figurnov,
  Olaf Ronneberger, Kathryn Tunyasuvunakool, Russ Bates, Augustin Ž{\'i}dek,
  Anna Potapenko, Alex Bridgland, Clemens Meyer, Simon A~A Kohl, Andy Ballard,
  Andrew Cowie, Bernardino Romera-Paredes, Stanislav Nikolov, Rishub Jain,
  Jonas Adler, Trevor Back, Stig Petersen, David Reiman, Ellen Clancy, Michal
  Zielinski, Martin Steinegger, Michalina Pacholska, Tamas Berghammer,
  Sebastian Bodenstein, David Silver, Oriol Vinyals, Andrew~W. Senior, Koray
  Kavukcuoglu, Pushmeet Kohli, and Demis Hassabis.
\newblock Highly accurate protein structure prediction with alphafold.
\newblock \emph{Nature}, 596:\penalty0 583 -- 589, 2021.
\newblock URL \url{https://api.semanticscholar.org/CorpusID:235959867}.

\bibitem[Devlin et~al.(2019)Devlin, Chang, Lee, and
  Toutanova]{Devlin2019BERTPO}
Jacob Devlin, Ming-Wei Chang, Kenton Lee, and Kristina Toutanova.
\newblock Bert: Pre-training of deep bidirectional transformers for language
  understanding.
\newblock In \emph{North American Chapter of the Association for Computational
  Linguistics}, 2019.
\newblock URL \url{https://api.semanticscholar.org/CorpusID:52967399}.

\bibitem[Kaplan et~al.(2020)Kaplan, McCandlish, Henighan, Brown, Chess, Child,
  Gray, Radford, Wu, and Amodei]{Kaplan2020ScalingLF}
Jared Kaplan, Sam McCandlish, Tom Henighan, Tom~B. Brown, Benjamin Chess, Rewon
  Child, Scott Gray, Alec Radford, Jeff Wu, and Dario Amodei.
\newblock Scaling laws for neural language models.
\newblock \emph{ArXiv}, abs/2001.08361, 2020.
\newblock URL \url{https://api.semanticscholar.org/CorpusID:210861095}.

\bibitem[Mishchenko et~al.(2007)]{mishchenko2007abc}
Alan Mishchenko et~al.
\newblock Abc: A system for sequential synthesis and verification.
\newblock \emph{URL http://www. eecs. berkeley. edu/alanmi/abc}, 17, 2007.

\bibitem[Belcak and Wattenhofer(2022)]{Belcak2022NeuralComb}
Peter Belcak and Roger Wattenhofer.
\newblock Neural combinatorial logic circuit synthesis from input-output
  examples.
\newblock \emph{CoRR}, abs/2210.16606, 2022.
\newblock \doi{10.48550/ARXIV.2210.16606}.
\newblock URL \url{https://doi.org/10.48550/arXiv.2210.16606}.

\bibitem[Zimmer et~al.(2023)Zimmer, Feng, Glanois, JIANG, Zhang, Weng, Li, HAO,
  and Liu]{zimmer2023differentiable}
Matthieu Zimmer, Xuening Feng, Claire Glanois, Zhaohui JIANG, Jianyi Zhang,
  Paul Weng, Dong Li, Jianye HAO, and Wulong Liu.
\newblock Differentiable logic machines.
\newblock \emph{Transactions on Machine Learning Research}, 2023.
\newblock ISSN 2835-8856.
\newblock URL \url{https://openreview.net/forum?id=mXfkKtu5JA}.

\bibitem[Hillier et~al.(2023)Hillier, Vũ, Mankowitz, Calandriello, Leurent,
  Rotival, Lobov, Mahajan, Gelmi, and Antropova]{hillier2023learning}
Adam Hillier, Ngân~(NV) Vũ, Daniel~J. Mankowitz, Daniele Calandriello,
  Edouard Leurent, Georges Rotival, Ivan Lobov, Kshiteej Mahajan, Marco Gelmi,
  and Natasha Antropova.
\newblock Learning to design efficient logic circuits, 2023.
\newblock URL \url{https://cassyni.com/events/S2LPTWZeMh9TGcLJe5jpqK}.
\newblock NANDA Workshop 2023.

\bibitem[Roy et~al.(2021)Roy, Raiman, Kant, Elkin, Kirby, Siu, Oberman, Godil,
  and Catanzaro]{Roy2021PrefixRL}
Rajarshi Roy, Jonathan Raiman, Neel Kant, Ilyas Elkin, Robert Kirby, Michael
  Siu, Stuart Oberman, Saad Godil, and Bryan Catanzaro.
\newblock Prefixrl: Optimization of parallel prefix circuits using deep
  reinforcement learning.
\newblock In \emph{2021 58th ACM/IEEE Design Automation Conference (DAC)},
  pages 853--858, 2021.
\newblock \doi{10.1109/DAC18074.2021.9586094}.

\bibitem[Li et~al.(2024{\natexlab{b}})Li, Shitole, Chien, Man, Wang, Srinivas,
  Zhang, Krishna, and Li]{li2024layerdag}
Mufei Li, Viraj Shitole, Eli Chien, Changhai Man, Zhaodong Wang, Srinivas, Ying
  Zhang, Tushar Krishna, and Pan Li.
\newblock Layer{DAG}: A layerwise autoregressive diffusion model of directed
  acyclic graphs for system.
\newblock In \emph{Machine Learning for Computer Architecture and Systems
  2024}, 2024{\natexlab{b}}.
\newblock URL \url{https://openreview.net/forum?id=IsarrieeQA}.

\bibitem[Amar{\'u} et~al.(2015)Amar{\'u}, Gaillardon, and
  De~Micheli]{EPFL-lsils}
Luca Amar{\'u}, Pierre-Emmanuel Gaillardon, and Giovanni De~Micheli.
\newblock The epfl combinational benchmark suite.
\newblock In \emph{Proceedings of the 24th International Workshop on Logic \&
  Synthesis (IWLS)}, number CONF, 2015.

\bibitem[{Meta Llama team}(2024)]{meta2024llama3}
{Meta Llama team}.
\newblock Introducing meta llama 3: The most capable openly available llm to
  date.
\newblock \emph{Meta AI Blog}, 2024.
\newblock URL \url{https://ai.meta.com/blog/meta-llama-3/}.

\bibitem[Karnaugh(1953)]{Karnaugh1953TheMM}
Maurice Karnaugh.
\newblock The map method for synthesis of combinational logic circuits.
\newblock \emph{Transactions of the American Institute of Electrical Engineers,
  Part I: Communication and Electronics}, 72:\penalty0 593--599, 1953.
\newblock URL \url{https://api.semanticscholar.org/CorpusID:51636736}.

\bibitem[Quine(1952)]{Quine1952}
W.~V. Quine.
\newblock The problem of simplifying truth functions.
\newblock \emph{The American Mathematical Monthly}, 59\penalty0 (8):\penalty0
  521--531, 1952.
\newblock ISSN 00029890, 19300972.
\newblock URL \url{http://www.jstor.org/stable/2308219}.

\bibitem[Quine(1955)]{Quine1955}
W.~V. Quine.
\newblock A way to simplify truth functions.
\newblock \emph{The American Mathematical Monthly}, 62\penalty0 (9):\penalty0
  627--631, 1955.
\newblock ISSN 00029890, 19300972.
\newblock URL \url{http://www.jstor.org/stable/2307285}.

\bibitem[McCluskey~Jr.(1956)]{McCluskey1956}
E.~J. McCluskey~Jr.
\newblock Minimization of boolean functions.
\newblock \emph{Bell System Technical Journal}, 35\penalty0 (6):\penalty0
  1417--1444, 1956.
\newblock \doi{https://doi.org/10.1002/j.1538-7305.1956.tb03835.x}.
\newblock URL
  \url{https://onlinelibrary.wiley.com/doi/abs/10.1002/j.1538-7305.1956.tb03835.x}.

\bibitem[Rudell and Sangiovanni-Vincentelli(1987)]{Rudell1987ESPRESSO}
R.L. Rudell and A.~Sangiovanni-Vincentelli.
\newblock Multiple-valued minimization for pla optimization.
\newblock \emph{IEEE Transactions on Computer-Aided Design of Integrated
  Circuits and Systems}, 6\penalty0 (5):\penalty0 727--750, 1987.
\newblock \doi{10.1109/TCAD.1987.1270318}.

\bibitem[Mishchenko et~al.(2011)Mishchenko, Brayton, Jang, and
  Kravets]{Mishchenko2011balance}
Alan Mishchenko, Robert Brayton, Stephen Jang, and Victor Kravets.
\newblock Delay optimization using sop balancing.
\newblock In \emph{2011 IEEE/ACM International Conference on Computer-Aided
  Design (ICCAD)}, pages 375--382, 2011.
\newblock \doi{10.1109/ICCAD.2011.6105357}.

\bibitem[Darringer et~al.(1981)Darringer, Joyner, Berman, and
  Trevillyan]{Darringer1981}
John~A. Darringer, William~H. Joyner, C.~Leonard Berman, and Louise Trevillyan.
\newblock Logic synthesis through local transformations.
\newblock \emph{IBM Journal of Research and Development}, 25\penalty0
  (4):\penalty0 272--280, 1981.
\newblock \doi{10.1147/rd.254.0272}.

\bibitem[Mishchenko et~al.(2006)Mishchenko, Chatterjee, and
  Brayton]{Mishchenko2006rewrite}
Alan Mishchenko, Satrajit Chatterjee, and Robert Brayton.
\newblock Dag-aware aig rewriting a fresh look at combinational logic
  synthesis.
\newblock In \emph{Proceedings of the 43rd Annual Design Automation
  Conference}, DAC '06, page 532–535, New York, NY, USA, 2006. Association
  for Computing Machinery.
\newblock ISBN 1595933816.
\newblock \doi{10.1145/1146909.1147048}.
\newblock URL \url{https://doi.org/10.1145/1146909.1147048}.

\bibitem[Riener et~al.(2019)Riener, Testa, Haaswijk, Mishchenko, Amarù,
  Micheli, and Soeken]{Riener2019GenericLS}
Heinz Riener, Eleonora Testa, Winston Haaswijk, Alan Mishchenko, Luca Amarù,
  Giovanni~De Micheli, and Mathias Soeken.
\newblock Scalable generic logic synthesis: One approach to rule them all.
\newblock In \emph{2019 56th ACM/IEEE Design Automation Conference (DAC)},
  pages 1--6, 2019.

\bibitem[Grosnit et~al.(2022)Grosnit, Malherbe, Tutunov, Wan, Wang, and
  Ammar]{grosnit2022boils}
Antoine Grosnit, Cedric Malherbe, Rasul Tutunov, Xingchen Wan, Jun Wang, and
  Haitham~Bou Ammar.
\newblock Boils: Bayesian optimisation for logic synthesis.
\newblock In \emph{2022 Design, Automation \& Test in Europe Conference \&
  Exhibition (DATE)}, pages 1193--1196. IEEE, 2022.

\bibitem[Feng et~al.(2022)Feng, Lyu, Chen, Ye, jie Yuan, and
  Hao]{Feng2022BatchSB}
Chang Feng, Wenlong Lyu, Zhitang Chen, Junjie Ye, Min jie Yuan, and Jianye Hao.
\newblock Batch sequential black-box optimization with embedding alignment
  cells for logic synthesis.
\newblock \emph{2022 IEEE/ACM International Conference On Computer Aided Design
  (ICCAD)}, pages 1--9, 2022.
\newblock URL \url{https://api.semanticscholar.org/CorpusID:254927570}.

\bibitem[Hosny et~al.(2020)Hosny, Hashemi, Shalan, and Reda]{Hosny2020DRiLLS}
Abdelrahman Hosny, Soheil Hashemi, Mohamed Shalan, and Sherief Reda.
\newblock {DRiLLS: Deep Reinforcement Learning for Logic Synthesis}.
\newblock \emph{2020 25th Asia and South Pacific Design Automation Conference
  (ASP-DAC)}, pages 581--586, September 2020.
\newblock \doi{10.1109/ASP-DAC47756.2020.9045559}.

\bibitem[Zhou and Anderson(2023)]{Zhou2023RLForest}
Guanglei Zhou and Jason~H. Anderson.
\newblock Area-driven fpga logic synthesis using reinforcement learning.
\newblock In \emph{2023 28th Asia and South Pacific Design Automation
  Conference (ASP-DAC)}, pages 159--165, 2023.

\bibitem[Qian et~al.(2024)Qian, Zhou, Zhou, and Wang]{Qian2024RL}
Yu~Qian, Xuegong Zhou, Hao Zhou, and Lingli Wang.
\newblock An efficient reinforcement learning based framework for exploring
  logic synthesis.
\newblock \emph{ACM Trans. Des. Autom. Electron. Syst.}, 29\penalty0 (2), jan
  2024.
\newblock ISSN 1084-4309.
\newblock \doi{10.1145/3632174}.
\newblock URL \url{https://doi.org/10.1145/3632174}.

\bibitem[Haaswijk et~al.(2018)Haaswijk, Collins, Seguin, Soeken, Kaplan,
  Süsstrunk, and De~Micheli]{Haaswijk2018GCN-RL}
Winston Haaswijk, Edo Collins, Benoit Seguin, Mathias Soeken, Frédéric
  Kaplan, Sabine Süsstrunk, and Giovanni De~Micheli.
\newblock Deep learning for logic optimization algorithms.
\newblock In \emph{2018 IEEE International Symposium on Circuits and Systems
  (ISCAS)}, pages 1--4, 2018.
\newblock \doi{10.1109/ISCAS.2018.8351885}.

\bibitem[Peruvemba et~al.(2021)Peruvemba, Rai, Ahuja, and
  Kumar]{Peruvemba2021GCN4RuntimeConstrRL}
Yasasvi~V. Peruvemba, Shubham Rai, Kapil Ahuja, and Akash Kumar.
\newblock Rl-guided runtime-constrained heuristic exploration for logic
  synthesis.
\newblock In \emph{2021 IEEE/ACM International Conference On Computer Aided
  Design (ICCAD)}, page 1–9. IEEE Press, 2021.
\newblock \doi{10.1109/ICCAD51958.2021.9643530}.
\newblock URL \url{https://doi.org/10.1109/ICCAD51958.2021.9643530}.

\bibitem[Zhu et~al.(2020)Zhu, Liu, Chen, Zhao, and Pan]{Zhu2020RLGCN}
Keren Zhu, Mingjie Liu, Hao Chen, Zheng Zhao, and David~Z. Pan.
\newblock Exploring logic optimizations with reinforcement learning and graph
  convolutional network.
\newblock In \emph{2020 ACM/IEEE 2nd Workshop on Machine Learning for CAD
  (MLCAD)}, pages 145--150, 2020.
\newblock \doi{10.1145/3380446.3430622}.

\bibitem[Basak~Chowdhury et~al.(2023)Basak~Chowdhury, Tan, Carey, Jain, Karri,
  and Garg]{Basak2023Bulls-Eye}
Animesh Basak~Chowdhury, Benjamin Tan, Ryan Carey, Tushit Jain, Ramesh Karri,
  and Siddharth Garg.
\newblock Bulls-eye: Active few-shot learning guided logic synthesis.
\newblock \emph{IEEE Transactions on Computer-Aided Design of Integrated
  Circuits and Systems}, 42\penalty0 (8):\penalty0 2580--2590, 2023.
\newblock \doi{10.1109/TCAD.2022.3226668}.

\bibitem[Yu et~al.(2018)Yu, Xiao, and De~Micheli]{Yu2018CNN4LS}
Cunxi Yu, Houping Xiao, and Giovanni De~Micheli.
\newblock Developing synthesis flows without human knowledge.
\newblock In \emph{Proceedings of the 55th Annual Design Automation
  Conference}, DAC '18, New York, NY, USA, 2018. Association for Computing
  Machinery.
\newblock ISBN 9781450357005.
\newblock \doi{10.1145/3195970.3196026}.
\newblock URL \url{https://doi.org/10.1145/3195970.3196026}.

\bibitem[Yu and Zhou(2020)]{Yu2020LSTM4LS}
Cunxi Yu and Wang Zhou.
\newblock Decision making in synthesis cross technologies using lstms and
  transfer learning.
\newblock In \emph{2020 ACM/IEEE 2nd Workshop on Machine Learning for CAD
  (MLCAD)}, pages 55--60, 2020.
\newblock \doi{10.1145/3380446.3430638}.

\bibitem[Wu et~al.(2022)Wu, Lee, Xie, and Hao]{Wu2022LOSTIN}
Nan Wu, Jiwon Lee, Yuan Xie, and Cong Hao.
\newblock Lostin: Logic optimization via spatio-temporal information with
  hybrid graph models.
\newblock In \emph{2022 IEEE 33rd International Conference on
  Application-specific Systems, Architectures and Processors (ASAP)}, pages
  11--18, 2022.
\newblock \doi{10.1109/ASAP54787.2022.00013}.

\bibitem[Bou et~al.(2023)Bou, Bettini, Dittert, Kumar, Sodhani, Yang,
  Fabritiis, and Moens]{bou2023torchrl}
Albert Bou, Matteo Bettini, Sebastian Dittert, Vikash Kumar, Shagun Sodhani,
  Xiaomeng Yang, Gianni~De Fabritiis, and Vincent Moens.
\newblock Torchrl: A data-driven decision-making library for pytorch, 2023.

\end{thebibliography}

\newpage

\appendix

\section{Notations}
Table \ref{tab:notation} contains a condensed summary of the notation introduced throughout the paper.  
\begin{table}[!ht]
    \caption{List of symbols and notations used in the paper.}
    \begin{center}
    \begin{tabular}{c|l}
    \textbf{Symbol} & \textbf{Meaning} \\
    \hline
    $n$ & Number of inputs in the AIG \\
    $I_j$ & Input node $j$ in the AIG \\
    $O$ & Output node for the AIG \\
    $\wedge_i$ & AND-node $i$ for the AIG \\
    $\Target{}$ & Target truth table \\
    $s$ & State \\
    $a$ & Action \\
    $\mathcal{T}$ & Set of truth tables in the AIG \\
    $\mathcal{A}$ & Set of actions \\
    $N$ & Current number of nodes in the AIG \\
    $N_{\text{max}}$ & Max number of nodes allowed in the AIG \OurMethod{} generates \\
    $\epsilon$ & Building type of an AND-node with respect to its two fanins (Table~\ref{tab:action_edge_types}) \\
    $\mathbf{A}$ & Sparse 3-dimensional tensor accumulating all the target actions \\
    $\mathbb{S}_k$ & Set of permutations of $\{1..k\}$ \\
    $\sigma(\cdot)$ & Random row permutation function \\
    $Q(s,a)$ & Discovered Q-value for a state $s$ and action $a$ \\
    $Q(s,a;\theta)$ & Predicted expected Q-value for a state $s$ and action $a$ \\
    $P(s,a; \theta)$ & Predicted action probability distribution \\
    $N(s,a)$ & Visit count of action $a$ and state $s$ \\
    $b, c$ & Parameter balancing exploration and exploitation in $\operatorname{PUCT}$ \\
    \end{tabular}
    \label{tab:notation}
    \end{center}
\end{table}

\section{Data collection}

\subsection{EPFL benchmarks}

Tables \ref{tab:arithmetic_epfl_datasets} and \ref{tab:ranodmcontrol_epfl_datasets}, contain detailed information about the arithmetic and random control circuits in the EPFL benchmarks~\citep{EPFL-lsils}, respectively. The circuits have been mapped from behavioral descriptions into logic gates and are intentionally suboptimal for scientific purposes. 
Arithmetic circuits, as their name hints, are combinatorial AIGs  representing an arithmetic operation such as square root, logarithm, etc., while the set of random control circuits consists of controller  circuits.
\begin{table}[h!]
    \caption{Arithmetic circuits in the EPFL benchmark suite and their statistics}
    \begin{center}
    \begin{tabular}{|c|c|c|c|c|}
    \hline
    \textbf{Circuit Name} & \textbf{\# Inputs} & \textbf{\# Outputs} & \textbf{\# AND-nodes} & \textbf{Levels} \\
    \hline
    \hline
    Adder & 256 & 129 & 1020 & 255 \\
    \hline
    Barrel Shifter & 135 & 128 & 3336 & 12 \\
    \hline
    Divisor & 128 & 128 & 44762 & 4470 \\
    \hline
    Hypotenuse & 256 & 128 & 214335 & 24801 \\
    \hline
    Log2 & 32 & 32 & 32060 & 444 \\
    \hline
    Max & 512 & 130 & 2865 & 287 \\
    \hline
    Multiplier & 128 & 128 & 27062 & 274 \\
    \hline
    Sine & 24 & 25 & 5416 & 225 \\
    \hline
    Square-root & 128 & 64 & 24618 & 5058 \\
    \hline
    Square & 64 & 128 & 18484 & 250 \\
    \hline
    \hline
    \textbf{Average:} & 166 & 102 & 37396 & 3608 \\
    \hline
    \end{tabular}
    \label{tab:arithmetic_epfl_datasets}
    \end{center}
\end{table}

\begin{table}[h!]
    \caption{Random/Control circuits in the EPFL benchmark suite and their statistics}
    \begin{center}
    \begin{tabular}{|c|c|c|c|c|}
    \hline
    \textbf{Circuit Name} & \textbf{\# Inputs} & \textbf{\# Outputs} & \textbf{\# AND-nodes} & \textbf{Levels} \\
    \hline
    \hline
    Round-Robin Arbiter & 256 & 129 & 11839 & 87 \\
    \hline
    Alu Control Unit & 7 & 26 & 174 & 10 \\
    \hline
    Coding-Cavlc & 19 & 11 & 693 & 16 \\
    \hline
    Decoder & 8 & 128 & 304 & 3 \\
    \hline
    i2c Controller & 147 & 142 & 1342 & 20 \\
    \hline
    Int to Float Converter & 11 & 7 & 260 & 16 \\
    \hline
    Memory Controller & 1204 & 1231 & 46836 & 114 \\
    \hline
    Priority Encoder & 128 & 8 & 978 & 250 \\
    \hline
    Lookahead XY Router & 60 & 30 & 257 & 54 \\
    \hline
    Voter & 1001 & 1 & 13758 & 70 \\
    \hline
    \hline
    \textbf{Average:} & 284 & 171 & 7644 & 64 \\
    \hline
    \end{tabular}
    \label{tab:ranodmcontrol_epfl_datasets}
    \end{center}
\end{table}

\subsection{Cut extraction}\label{sec:cut-extraction}
To extract a cut from a given node with a target number of inputs $n$, we designate the given node as the root and of the cut and its two parents as the initial leaf set.
We then iteratively remove a random node from the leaf set and add its parents to the leaf set while maintaining the leaf property.
This process continues until the leaf set contains $n$ nodes.
Finally, we create an AIG out of the visited nodes within the cut where we mark the leaf set as the inputs and the root node as the output.
We provide the outline of the procedure in Algorithm~\ref{alg:aig_extraction}.

\begin{algorithm}[!hb]
    \caption{Cut Extraction} \label{alg:aig_extraction}
    \begin{algorithmic}[1]
        \REQUIRE Root node: $r$, number of cut inputs: $n$
        \STATE leaf\_set = \{left\_parent($r$), right\_parent($r$)\}
        \STATE node\_set = \{$r$\}
        \WHILE{size(leaf\_set) < $n$}
        \STATE node = leaf\_set.random\_pop() 
        \STATE node\_set.insert(node)
        \STATE leaf\_set.insert(left\_parent(node)) \& leaf\_set.insert(right\_parent(node))
        \STATE Ensure leaf property in leaf\_set
        \ENDWHILE
        \STATE Construct AIG from leaf\_set and node\_set
    \end{algorithmic}
\end{algorithm}

We can modify this algorithm to extract additional cuts per node by repeating the process until we find a cut with $n-1$ leaf nodes.
For this cut, instead of randomly expanding a leaf node, we create $n-1$ copies of the cut and expand each leaf node individually, storing the resulting cuts.
In practice, we actually employ Algorithm \ref{alg:multi_aig_extraction} to extract AIGs from the EPFL circuits.
Although Algorithm~\ref{alg:aig_extraction} conveys the core idea of AIG extraction, the following algorithm is more effective from an engineering standpoint, as it allows to extract more cuts from the same node.

The revised algorithm takes two inputs: $n$, the desired number of input nodes, and the root node.
It initializes the leaf set by adding the parents of the root node.
It then iteratively removes a random node from the leaf set and expands it using Algorithm \ref{alg:cut_expansion}.
This process continues until the cut contains $n-1$ leaf nodes, at which point we create $n-1$ copies of the current cut state and expand each leaf node to generate $n$ unique cuts.
For brevity, we omitted the details of ensuring the leaf property after node expansion in the main text, which is addressed in Algorithm \ref{alg:preseve_leaf_prop}.
This algorithm ensures that no node in the leaf set has a parent also in this set, which would violate the leaf property.
If a parent is already in the leaf set, we can remove the node from the set and add the other parent that is not yet in the leaf set.

\begin{algorithm}[h!]
    \caption{Multi-Cut Extraction} \label{alg:multi_aig_extraction}
    \begin{algorithmic}[1]
        \REQUIRE Root node: $r$, Number of cut inputs: $n$
        \STATE leaf\_set = \{left\_parent($r$), right\_parent($r$)\}
        \STATE node\_set = \{$r$\}
        \WHILE{size(leaf\_set) < $n-1$}
        \STATE node = leaf\_set.random\_pop()
        \STATE Cut\_Expansion(node, leaf\_set, node\_set)
        \ENDWHILE
        \FOR{leaf in leaf\_set}
        \STATE copy\_leaf\_set = leaf\_set.copy() \& copy\_node\_set = node\_set.copy()
        \STATE copy\_leaf\_set.delete(leaf)
        \STATE Cut\_Expansion(leaf, copy\_leaf\_set, copy\_node\_set)
        \STATE Construct AIG from copy\_leaf\_set and copy\_node\_set
        \ENDFOR
        
    \end{algorithmic}
\end{algorithm}

\begin{algorithm}[h!]
    \caption{Cut Expansion} \label{alg:cut_expansion}
    \begin{algorithmic}[1]
        \REQUIRE Node to expand: node, Current leaf nodes: leaf\_set, Current nodes in cut: node\_set
        \STATE node\_set.insert(node)
        \STATE leaf\_set.insert(left\_parent(node)) \& leaf\_set.insert(right\_parent(node))
        \STATE preserve\_leaf\_property(leaf\_set, node\_set)
    \end{algorithmic}
\end{algorithm}

\begin{algorithm}[h!]
    \caption{Preserve Leaf Property} \label{alg:preseve_leaf_prop}
    \begin{algorithmic}[1]
        \REQUIRE Current leaf nodes: leaf\_set, Current nodes in cut: node\_set
        \FOR{leaf in leaf\_set}
        \IF{left\_parent(leaf) in leaf\_set}
        \STATE leaf\_set.delete(leaf) \& node\_set.insert(leaf)
        \STATE leaf\_set.insert(right\_parent(node))
        \ELSIF{right\_parent(leaf) in leaf\_set}
        \STATE leaf\_set.delete(leaf) \& node\_set.insert(leaf)
        \STATE leaf\_set.insert(left\_parent(node))
        \ENDIF
        \ENDFOR
    \end{algorithmic}
\end{algorithm}


\section{Additional related works} \label{app:add-related-works}
\subsection{Heuristics for AIG Generation and Optimization} \label{app:heuristics}
As the inference of CNFs using exact SAT solvers often lead to exponentially large expressions,  various heuristics
such as Karnaugh maps 
\citep{Karnaugh1953TheMM}, or Quine-McCluskey methods
~\citep{Quine1952, Quine1955, McCluskey1956}),
and algorithms~\citep{Rudell1987ESPRESSO} have been designed to obtain more compact expressions or circuits.
Further efforts
accompanying the  rise in chip demand  
led to the development of widely used logic synthesis libraries that implement equivalence-preserving Boolean network operators. 
The open-source library \texttt{ABC}~\citep{mishchenko2007abc}) notably comprises dozens of logic graph operators aiming at reducing a network size or depth \citep{Mishchenko2011balance, Mishchenko2006ScalableLS}.
Interestingly, some important operators such as \texttt{resub} or \texttt{rewrite} \citep{Darringer1981, Mishchenko2006rewrite} acts on the subject graph through a series of local modifications involving small single-output AIGs.
Besides, applying a single operator on a logic network is suboptimal compared to applying several operators sequentially, though finding the best sequence is also a hard problem~\citep{Riener2019GenericLS}.

\subsection{Machine Learning for Logic Synthesis}
Many ML approaches have been explored to tackle the operator flow optimization progress.
Some stateless optimization methods, such as Bayesian optimization~\citep{grosnit2022boils, Feng2022BatchSB}, search for the best flow without considering  the subject graph specificities. 
Alternatively, state-based methods formulate the operator sequence optimization as a Reinforcement learning problem, and train policies on selected features of the logic network.
While some works use high-level statistics of the subject graph (e.g., its number of nodes)~\citep{Hosny2020DRiLLS, Zhou2023RLForest, Qian2024RL}, others rely on tailored graph convolutional networks (GCN) to extract richer features at the cost of a longer training time~\citep{Haaswijk2018GCN-RL,Peruvemba2021GCN4RuntimeConstrRL, Zhu2020RLGCN, Basak2023Bulls-Eye}. 
Similarly,  standard deep network architectures, such as CNNs~\citep{Yu2018CNN4LS},  LSTMs~\citep{Yu2020LSTM4LS}, or GCNs~\citep{Wu2022LOSTIN} have been trained to predict the quality of a logic synthesis flow in a supervised way.  
Contrary to these works, we target AIG generation itself and not operator optimization. 

\section{Training Parameters and Implementation}
We implement \OurMethod{} with PyTorch and TorchRL~\citep{bou2023torchrl}.
During pre-training, We use a cosine annealing with warm restarts learning rate scheduler with a starting learning rate of $1\times 10^{-3}$ and a batch size of 1024 for 250 epochs. During fine-tuning we use a batch size of 128, a replay buffer with capacity of 1M and sync the parameters every 500 training steps. 

\end{document}